\documentclass{article}
\usepackage[most]{tcolorbox}
\usepackage{fvextra}
\tcbuselibrary{skins,breakable}
\newenvironment{promptbox}[1][]
  {\begin{tcolorbox}[
    enhanced,
    breakable,
    title=#1,
    colback=gray!5!white,
    colframe=gray!80!black,
    width=\linewidth,
    boxsep=2pt,
    left=2pt,
    right=2pt,
    top=4pt,
    bottom=4pt,
    fonttitle=\fontsize{8.5pt}{10pt}\selectfont\bfseries,
  ]}
  {\end{tcolorbox}}

\RecustomVerbatimEnvironment{verbatim}{Verbatim}{
  fontsize=\fontsize{8pt}{10pt}\selectfont,
  breaklines=true,
  breakanywhere=true,
  breaksymbolleft={},
  breakindent=0pt,
  xleftmargin=0pt,
  xrightmargin=0pt,
  tabsize=2
}

\usepackage[most]{tcolorbox}
\tcbuselibrary{listings, skins}
\usepackage{listings}
\usepackage{xcolor}

\PassOptionsToPackage{numbers, sort}{natbib}
    
\usepackage[main, final]{neurips_2026}
\usepackage{pifont}

\usepackage[utf8]{inputenc} 
\usepackage[T1]{fontenc}    
\usepackage{url}            
\usepackage{booktabs}       
\usepackage{amsfonts}       
\usepackage{nicefrac}       
\usepackage{microtype}      
\usepackage{xcolor}         
\usepackage{colortbl}
\usepackage[dvipsnames,svgnames,table]{xcolor}
\usepackage[linktoc=all]{hyperref} 
\usepackage{algorithm}
\usepackage{algpseudocode}
\usepackage{graphicx}
\usepackage{mwe}  
\usepackage{multirow}
\usepackage{subcaption}
\usepackage{amssymb}
\usepackage{amsmath}
\usepackage{fvextra}
\usepackage{makecell}

\hypersetup{
    colorlinks=true,
    linkcolor=RoyalBlue,
    citecolor=RoyalBlue,
    filecolor=cyan,
    urlcolor=RoyalBlue
}

\definecolor{custom_green}{rgb}{0.0, 0.5, 0.0}
\definecolor{custom_orange}{RGB}{255,140,0}
\definecolor{custom_red}{rgb}{1.0, 0.01, 0.24}
\newcommand{\cmark}{\color{custom_green}\ding{51}}%
\newcommand{\xmark}{\color{custom_red}\ding{55}}%
\newcommand{\tmark}{\textcolor{custom_orange}{$\pmb{\triangle}$}}

\usepackage{hyperref}
\usepackage{refcount}

\makeatletter
\def\aref@figure{figure}
\def\aref@subfigure{subfigure}
\def\aref@table{table}
\def\aref@subtable{subtable}

\newcommand{\aref}[1]{%
  \begingroup
  \edef\aref@anchor{\getrefbykeydefault{#1}{anchor}{section.0}}%
  \expandafter\aref@parse\aref@anchor.\@nil{#1}%
  \endgroup
}

\def\aref@parse#1.#2\@nil#3{%
  \def\aref@type{#1}%
  \ifx\aref@type\aref@figure
    \hyperref[#3]{Figure~\ref*{#3}}%
  \else\ifx\aref@type\aref@subfigure
    \hyperref[#3]{Figure~\ref*{#3}}%
  \else\ifx\aref@type\aref@table
    \hyperref[#3]{Table~\ref*{#3}}%
  \else\ifx\aref@type\aref@subtable
    \hyperref[#3]{Table~\ref*{#3}}%
  \else
    \edef\aref@num{\getrefnumber{#3}}%
    \expandafter\aref@secdispatch\aref@num\@nil{#3}%
  \fi\fi\fi\fi
}

\def\aref@secdispatch#1#2\@nil#3{%
  \ifnum`#1>`9\relax
    \hyperref[#3]{Appendix~\ref*{#3}}%
  \else
\hyperref[#3]{Section~\ref*{#3}}%
  \fi
}
\makeatother

\title{LLM-as-a-Tutor: Policy-Aware\\Prompt Adaptation for Non-Verifiable RL}

\author{%
Yujin Kim${}^{1}$\thanks{
Equal contribution.\,\,\,
${}^\dagger$Corresponding authors.
}
\,\,\,\, Namgyu Ho${}^{1*}$ \,\,\, Sangmin Hwang${}^{1*}$ \,\,\, Joonkee Kim${}^{2}$ \,\,\, Yongjin Yang${}^{3}$ \vspace{2pt}\\ 
\textbf{Sangmin Bae}${}^{1}$ \,\,\, \textbf{Seungone Kim}${}^{4}$ \,\,\, \textbf{Jaehun Jung}${}^{5}$ \,\,\, \textbf{Se-Young Yun}${}^{1\dagger}$ \,\,\, \textbf{Hwanjun Song}${}^{1,6\dagger}$ \vspace{2pt}\\
  ${}^{1}$KAIST \quad ${}^{2}$Upstage \quad ${}^{3}$University of Toronto \quad ${}^{4}$Carnegie Mellon University \\
  ${}^{5}$NVIDIA \quad ${}^{6}$Cluvion \vspace{2pt} \\
  \texttt{\{yujin399,itsnamgyu,sangmin.hwang\}@kaist.ac.kr} \\
  \url{https://github.com/bbang3/LLM-as-a-Tutor}
  }     

\begin{document}

\maketitle

\begin{abstract}
Reinforcement learning (RL) for non-verifiable instruction following increasingly relies on LLM judges with prompt-specific rubrics as reward signals.
While recent methods adapt these rubrics to the evolving policy during training, the training prompts themselves remain static, drawn from fixed corpora.
This static approach often results in a critical misalignment between prompt difficulty and policy capability, leaving the judge unable to recover a discriminative reward signal when prompts fail to elicit quality variance among rollouts.
To address this misalignment, we introduce \textit{LLM-as-a-Tutor}, a framework that extends the LLM's role from judge to tutor: a single model serves as an examiner that pairwise-compares policy rollouts to detect non-challenging prompts, and as a generator that appends atomic constraints to them.
This append-only design preserves the original prompt while adding constraints to increase difficulty in step with the policy’s capability, producing a self-calibrating training signal without external difficulty schedules.
On three complex instruction-following benchmarks, our method outperforms both policy-unaware baselines and prior policy-adaptive methods that adapt rubrics or rewrite prompts, on average and on four of five metrics, suggesting prompt adaptation as an underexplored axis of policy-awareness in non-verifiable RL for constraint-heavy instruction following.
\end{abstract}
\begin{figure}[h]
    \small
    \centering
    \includegraphics[width=\linewidth,keepaspectratio]{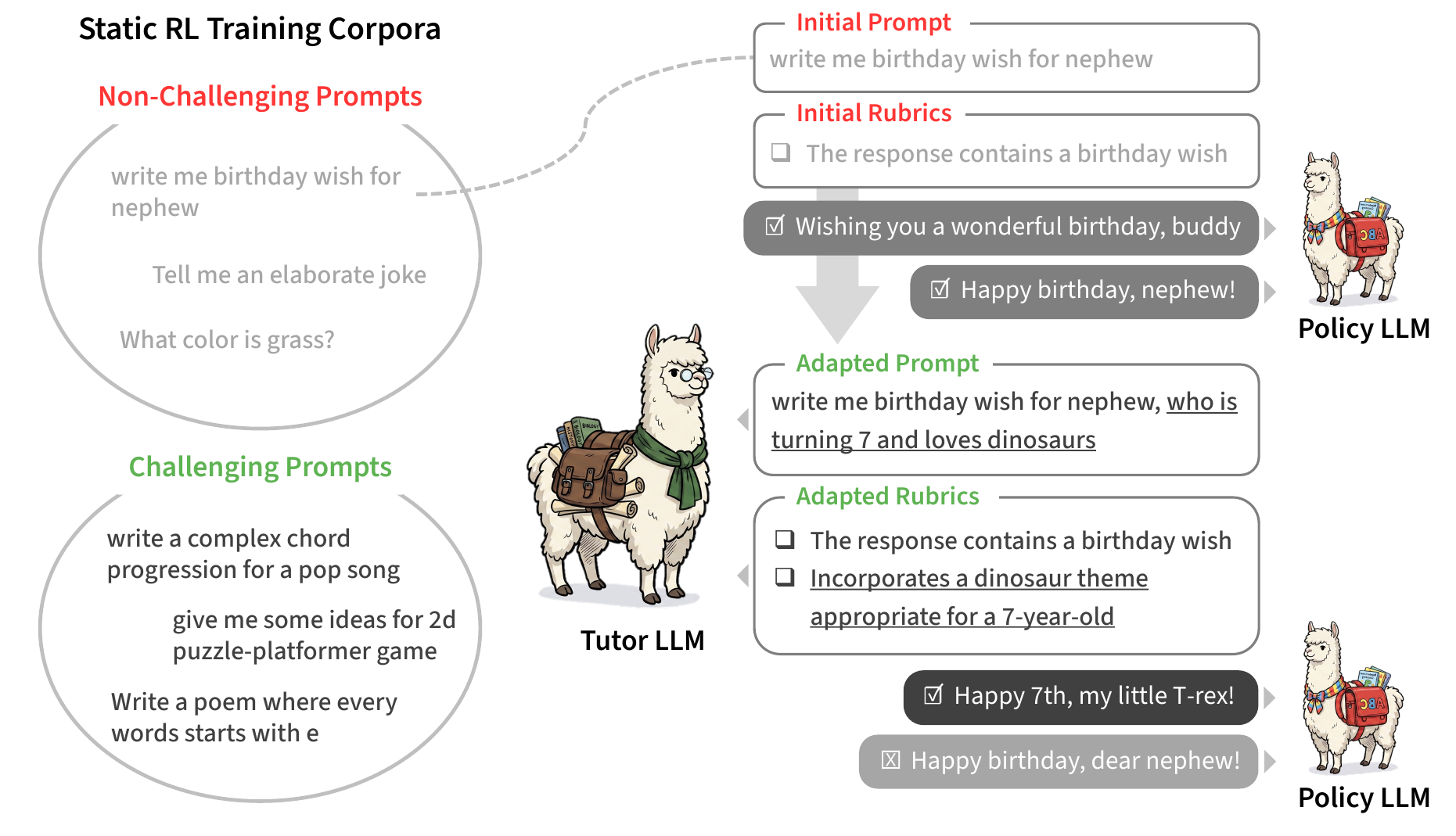}
    \caption{
    \textbf{Overview of LLM-as-a-Tutor: when the policy's responses to a prompt are indistinguishable in quality, a tutor LLM adds a constraint to make the prompt more challenging.}
    (Left) Static RL training corpora for non-verifiable tasks contain prompts that are non-challenging for the current policy and provide little learning signal.
    (Right) The tutor examines a pair of policy rollouts; if their quality is indistinguishable, the tutor appends a constraint and corresponding rubric criteria which turn the non-challenging prompt into a challenging one that elicits variance in rollouts.
    }
    \label{fig:method_concept}
\end{figure}

\section{Introduction}
\label{sec:introduction}

Reinforcement learning depends on the reward signal's ability to \textit{discriminate} among policy outputs of varying quality \cite{bae2026online, tzannetos2023proximal, sutton2018rl, williams1992reinforce, schulman2017ppo}.
In verifiable domains such as math and code, programmatic checkers provide this signal directly \cite{shao2024deepseekmath, guo2025deepseekr1} but open-ended instruction following has no analogous checker.
Bradley-Terry reward models \cite{christiano2017deep, stiennon2020learning, ouyang2022training} provide only coarse discrimination, leading to reward hacking and plateaued performance on complex benchmarks \cite{casper2023open, viswanathan2025checklists, he2025advancedif, qin2024infobench}.
Recent work conditions LLM judges on instance-specific rubrics \cite{viswanathan2025checklists, shao2025drtulu, gunjal2026rubrics}, recovering fine-grained discrimination and generalizing zero-shot to novel domains \cite{kim2024prometheus}.

The discriminative power of the LLM judge, however, comes with a precondition: the prompt itself must \textit{induce} rollouts that differ in quality.
We refer to such prompts as \textit{challenging} for the current policy.
When a prompt is too easy for the current policy, all rollouts succeed; when too hard, all rollouts fail. 
In either case, the rubric judge has nothing to discriminate, and the reward signal collapses regardless of judge quality.
This is a property of the policy-prompt pair, not of the prompt alone.
Yet existing pipelines for non-verifiable RL with LLM judges draw training prompts from static corpora \cite{gunjal2026rubrics, viswanathan2025checklists, shao2025drtulu}, leaving the supply of \textit{challenging} prompts to chance.

A natural response is to adapt prompts to the current policy so they reliably induce discriminative rollouts, a property we call \textit{policy-adaptiveness}.
Recent work in non-verifiable RL pursues this but does not fully exploit the discriminative capabilities of LLMs.
One line uses scalar reward-model scores \cite{ye2025evolving}, inheriting the coarse-discrimination problem that motivated rubric-equipped judges in the first place.
Another uses LLM judges with generic, prompt-agnostic rubrics \cite{kuba2025language}, which lack the fine-grained discrimination instance-specific rubrics provide \cite{gunjal2026rubrics, kim2024prometheus, liu2023geval}.

We propose \textit{LLM-as-a-Tutor}, a method that adapts prompts to the current policy by exploiting the discriminative capabilities of LLMs that prior work has left unused. Detecting whether a prompt is challenging for the current policy is a fundamentally pairwise question, namely whether two rollouts from the policy differ in quality. Pairwise judgment is the mode in which LLMs are consistently sharper and more human-aligned than pointwise scoring \cite{zheng2023judging, liu2024aligning, liusie2024comparative}. We extend the role of the LLM from judge to tutor, in which a single LLM plays two roles: an \textit{examiner} that compares rollout pairs to detect non-challenging prompts, and a \textit{generator} that augments them.

Concretely, the tutor operates as follows. Each seed prompt is paired with a base rubric, generated once by the tutor, that serves as the reward signal during training. At each iteration, the examiner samples a pair of rollouts under the current policy and assesses whether they differ in quality. If they do not, the generator appends an additional constraint to the prompt and corresponding criteria to the rubric in a single call, while leaving the seed prompt and base rubric otherwise intact.
In complex instruction following, difficulty is naturally instantiated as the number of requirements a model must simultaneously satisfy \cite{jiang2024followbench}, so adding constraints provides a way to increase difficulty in step with the policy's capability while preserving the original prompt.

%

We evaluate LLM-as-a-Tutor on three complex instruction-following benchmarks, FollowBench \cite{jiang2024followbench}, AdvancedIF \cite{he2025advancedif}, and InfoBench \cite{qin2024infobench}, using Qwen3-1.7B \cite{yang2025qwen3} as the policy and Qwen3-8B as the tutor.
Our method achieves the highest average score, outperforming both baselines that train on fixed prompts (base rubrics, WildChecklists \cite{viswanathan2025checklists}, policy-adaptive rubrics) and baselines that modify prompts (Evol-Instruct \cite{xu2023wizardlm}, EVA \cite{ye2025evolving}).
Notably, policy-adaptive rubrics also underperform on average, suggesting that for some prompts no rubric refinement can recover discriminative signal and the prompt itself must be changed.
Ablations confirm that the gains come from policy-aware adaptation: applying constraints unconditionally, at random, or against a stronger model's responses all underperform targeted appending, and appending outperforms rewriting prompts.
Analysis shows that the tutor's pairwise judgments correlate with empirical difficulty, that appended constraints amplify the discriminative signal carried by the base rubric, and that the modification rate rises monotonically with policy scale.

%


Our contributions are as follows:
\begin{itemize}
\item We identify \textit{policy-prompt mismatch} as a precondition that limits rubric-equipped reward judges in non-verifiable RL.
\item We propose \textit{LLM-as-a-Tutor}, which extends the LLM's role from judge to tutor with two mechanisms: a pairwise saturation trigger that detects non-challenging prompts via pairwise comparison of two policy rollouts \cite{zheng2023judging, liu2024aligning, liusie2024comparative}, and append-only constraint generation that augments them with atomic constraints grounded in those rollouts.
\item Our method outperforms both policy-unaware and policy-adaptive baselines on three instruction-following benchmarks in average score under two evaluation judges, suggesting prompt adaptation as an underexplored axis of policy-awareness in non-verifiable RL for constraint-heavy instruction following.
\end{itemize}

\section{Related work}
\label{sec:related}

\subsection{Reward variance as a learnability signal}
\label{related_variance}

A long-standing principle in RL is that learning requires reward variation across sampled trajectories: when all trajectories from a state receive identical returns, the policy gradient carries no signal about which action was preferable. This motivates curriculum methods that select tasks at the frontier of agent competence~\citep{tzannetos2023proximal, bengio2009curriculum}, and the same idea has recently been adapted to LLM RL by filtering prompts whose empirical pass rate is too high or too low~\citep{bae2026online} or by sampling prompts in proportion to per-prompt reward variance~\citep{jiang2025vcrl}. A parallel line of self-play methods~\citep{huang2025rzero, zhao2025absolute, liu2025spice} co-evolves task generators with solvers under moderate-difficulty objectives so that neither side of the success distribution collapses. All of these assume binary, verifiable rewards, whereas we target rollout-level discriminative variance under continuous, rubric-based rewards in non-verifiable instruction following.





\subsection{Rubric-based non-verifiable RL}
\label{related_rubric_rl}

Standard non-verifiable RL relies on scalar reward models trained on preference data, but these signals are coarse and brittle---prone to reward hacking, biased by surface features such as response length, and unstable under optimization~\citep{casper2023open}. Building on rubric-based fine-grained evaluation~\citep{kim2024prometheus, liu2023geval}, recent work instead repurposes instance-specific rubrics as reward signals for on-policy RL: RLCF~\citep{viswanathan2025checklists}, Rubrics as Rewards~\citep{gunjal2026rubrics}, and AdvancedIF~\citep{he2025advancedif} all train against per-prompt checklists scored by LLM judges, while OpenRubrics~\citep{liu2025openrubrics} tackles the complementary problem of scaling rubric construction itself. A subsequent direction makes the rubrics adaptive rather than fixed: DR Tulu~\citep{shao2025drtulu} maintains rubrics that co-evolve with the policy during training, continuously tailoring criteria to its current behavior so that the reward signal remains discriminative as the model improves.



\subsection{Policy-adaptive methods for RL}
\label{related_policy_adaptive}

A line of work adapts the training distribution to the policy's current capability during RL.
DR Tulu~\citep{shao2025drtulu} evolves rubrics rather than prompts, while EVA~\citep{ye2025evolving} and LSP~\citep{kuba2025language} evolve prompts but rely on scalar RMs or generic, prompt-agnostic rubrics, forgoing the fine-grained signal of instance-specific rubrics.
Our method extends policy-adaptive prompt evolving to the rubric-based setting, using instance-specific rubric judges with pairwise saturation checks~\citep{zheng2023judging} and appending atomic constraints to adapt prompt difficulty to the policy's evolving capabilities while preserving the original prompt.


\section{Method}
\label{sec:method}

\subsection{Preliminaries}\label{subsec:prelim}

\paragraph{Rubric-based reward.}
We consider RL for non-verifiable instruction-following tasks, where for each prompt $x \sim \mathcal{D}$, a prompt-specific rubric $R(x) = \{(r_k, w_k)\}_{k=1}^{K}$ defines a set of natural-language criteria $r_k$ with non-negative weights $w_k$ satisfying $\sum_{k=1}^{K} w_k = 1$.
We refer to $R(x)$ as the \emph{base rubrics}, since it captures broad quality criteria for $x$ independent of any particular policy.
The reward score $s$ for a response $y$ given a target prompt $x$ is computed as a weighted sum of per-criterion judgments from an LLM judge $\mathcal{J}$:
\begin{equation}
    s(x, y) = \sum_{k=1}^{K} w_k \cdot \mathcal{J}(y \mid x, r_k),
    \label{eq:rubric_reward}
\end{equation}
where $\mathcal{J}(y \mid x, r_k) \in [0, 100]$\footnote{The score range is determined by the judge prompt; we use $[0, 100]$ (refer to \aref{app:judge_prompt}). The absolute scale does not affect learning since GRPO normalizes rewards within each rollout group.}
 denotes the judge's score for how well $y$ satisfies criterion $r_k$. Following prior work~\citep{viswanathan2025checklists, shao2025drtulu}, $\mathcal{J}$ evaluates each criterion independently, producing a separate judgment per criterion. Training on a fixed prompt corpus with reward computed solely from the base rubrics $R(x)$ constitutes our baseline.

\paragraph{Policy optimization with GRPO.}
For each prompt $x$, we sample $G$ rollouts $\{y^{(i)}\}_{i=1}^{G}$ from the old policy $\pi_{\theta_{\text{old}}}$ and optimize the policy $\pi_\theta$ by maximizing the clipped surrogate objective~\cite{shao2024deepseekmath, schulman2017ppo}
\begin{equation}
\resizebox{0.945\linewidth}{!}{$
\displaystyle
\mathcal{J}_{\text{GRPO}}(\theta) = \mathbb{E}\!\left[ \frac{1}{G} \sum_{i=1}^{G} \frac{1}{|y^{(i)}|} \sum_{t=1}^{|y^{(i)}|} \min\!\left( \rho^{(i)}_{t}(\theta)\, A^{(i)},\, \mathrm{clip}(\rho^{(i)}_{t}(\theta), 1-\varepsilon, 1+\varepsilon)\, A^{(i)} \right) - \beta\, \mathbb{D}_{\text{KL}}[\pi_\theta \,\|\, \pi_{\text{ref}}] \right]
$}
\label{eq:grpo_objective}
\end{equation}
with per-token importance ratio $\rho^{(i)}_{t}(\theta) = \pi_{\theta}(y^{(i)}_{t} \mid x, y^{(i)}_{<t}) / \pi_{\theta_{\text{old}}}(y^{(i)}_{t} \mid x, y^{(i)}_{<t})$, clipping range $\varepsilon$, and KL coefficient $\beta$ regularizing toward a frozen reference policy $\pi_{\text{ref}}$. Following~\citet{shao2024deepseekmath}, the per-rollout advantage $A^{(i)}$ is broadcast to every token of $y^{(i)}$ and computed in a group-relative manner,
\begin{equation}
    A^{(i)} = \frac{s(x, y^{(i)}) - \mu(x)}{\sigma(x)},
    \label{eq:grpo_advantage}
\end{equation}
where $\mu(x)$ and $\sigma(x)$ are the mean and standard deviation of $\{s(x, y^{(i)})\}_{i=1}^{G}$ (with a small constant added to $\sigma(x)$ for numerical stability).  The advantage $A^{(i)}$ provides a meaningful learning signal only when the rollouts within a group exhibit sufficient spread in score.

\paragraph{Discriminative signal as a property of the prompt--policy pair.}
Whether a prompt induces rollouts of varying quality is a property of the prompt paired with the current policy, not of the prompt alone. When the pair is misaligned, the rollouts within a group offer little contrast for the advantage to exploit, regardless of the quality of $R(x)$ or $\mathcal{J}$.
Such misalignment arises naturally when prompts are drawn from static corpora~\citep{zhao2024wildchat} and tends to worsen as the policy improves under RL~\citep{bae2026online, ye2025evolving, jiang2025vcrl}.
Closing this gap requires adapting the prompts themselves to the policy's evolving capability.

\subsection{LLM-as-a-Tutor}

\paragraph{Pipeline overview.}
As shown in \aref{fig:method_concept}, \emph{LLM-as-a-Tutor} adapts seed prompts and their initial base rubrics
in step with the evolving capabilities of the policy model using a separate tutor LLM.
At the start of each epoch, the tutor examines each prompt
together with two rollouts sampled from the current policy.
If the prompt is judged non-discriminative, the tutor appends
an atomic constraint to the prompt and adds matching criteria
to its rubric; otherwise, the prompt--rubric pair remains
unchanged. The resulting prompt--rubric pairs are then used
to train the policy via rubric-based RL. 

Our method comprises three key components:
pairwise discriminativeness judgment, constraint-based prompt
adaptation, and iterative difficulty escalation. We detail the
design rationale for each below.


\paragraph{Pairwise discriminativeness judgment.}
The tutor decides whether $x$ is discriminative for $\pi_\theta$ by comparing two rollouts from $\pi_\theta$. We use a pairwise comparison rather than independent scoring since LLMs are consistently sharper and more human-aligned in pairwise comparisons than in pointwise scoring~\citep{zheng2023judging, liu2024aligning, liusie2024comparative}. Specifically, the tutor samples two rollouts $y^{(1)}, y^{(2)} \sim \pi_\theta(\cdot \mid x)$ and makes a binary judgment on whether they are \emph{indistinguishable in quality}; if so, $x$ is declared \emph{non-discriminative} for the current policy. The judgment considers whether the two rollouts agree in quality, in their overall approach, and in the absence of meaningful weaknesses. 
We verify in \aref{sec:analysis} that prompts judged non-discriminative indeed have higher mean reward and lower variance across the full rollout group than those judged discriminative.

\paragraph{Constraint-based prompt adaptation.}
When $x$ is non-discriminative, the tutor invokes its second role and produces an atomic constraint $c$ together with rubrics $R_c = \{(r_{c,j}, w_{c,j})\}_{j=1}^{m}$ scoring adherence to it.
The constraint $c$ is \emph{atomic} in the sense of imposing a single requirement along a dimension left unspecified by $x$.
To prevent conflicting or hallucinated constraints, the tutor sees the full accumulated prompt and writes a constraint that is a seamless continuation of it, factually grounded, and concrete and evaluable.
The constraint is appended to form the adapted prompt and rubric set,
\begin{equation}
    \tilde{x} = x \oplus c, \qquad \tilde{R}(x) = R(x) \cup R_c,
    \label{eq:adapted}
\end{equation}
with the weights of $R(x)$ and $R_c$ jointly renormalized to sum to one. Discriminative prompts are left untouched: $(\tilde{x}, \tilde{R}(x)) = (x, R(x))$. The reward and GRPO objective from \aref{subsec:prelim} are then applied unchanged with $(x, R(x))$ replaced by $(\tilde{x}, \tilde{R}(x))$, so that adaptation is the sole locus of change relative to the baseline.

This append-only design adds new constraints while preserving the original requirements: any response that satisfies $\tilde{x}$ must also satisfy $x$.
Strategies that rewrite the prompt offer no such guarantee, and we show in \aref{subsec:ablation_rewrite} that a rewriting variant underperforms appending.
Retaining $R(x)$ in $\tilde{R}(x)$ also keeps the original criteria of $x$ active in the reward. As $\tilde{x}$ accumulates constraints and the policy must satisfy them simultaneously, scoring well on $R(x)$ itself grows increasingly demanding, leaving the base rubrics to continue to carry discriminative signal across iterations (\aref{sec:analysis}).

\paragraph{Difficulty escalation across iterations.}
The tutor is invoked at the start of every training iteration, where an iteration denotes any chosen adaptation interval (e.g., every epoch, every $N$ train steps) so that prompts adapted in earlier iterations can be revisited as the policy improves. Let $x^{(t)}$ denote the prompt presented to the policy at iteration $t$, with $x^{(0)} = x$. If $x^{(t)}$ is non-discriminative for $\pi_{\theta_t}$, the tutor appends a new atomic constraint $c^{(t)}$ to form $x^{(t+1)} = x^{(t)} \oplus c^{(t)}$, with $\tilde{R}(x^{(t+1)})$ accumulating the corresponding rubrics; otherwise $x^{(t+1)} = x^{(t)}$.
The sequence $\{x^{(t)}\}$ is non-decreasing in difficulty for any policy, while the trigger for escalation is set by the current $\pi_{\theta_t}$.
Prompt difficulty thus escalates in step with the policy's improving capability, realizing the dynamic adaptation that a fixed prompt corpus cannot provide.

The complete training pipeline is described in \aref{app:pseudo_code}, and the tutor prompt templates are provided in \aref{app:tutor_prompt}.

\section{Experiments}
\label{sec:experiments}

\subsection{Experimental setup}
We train Qwen3-1.7B-Thinking~\citep{yang2025qwen3} as the policy, with Qwen3-8B-Thinking serving as both tutor and judge. We train for 3 epochs on randomly sampled 4K prompts from WildChat~\citep{zhao2024wildchat}, and we instantiate the tutor's adaptation interval as one full epoch. We evaluate on three instruction-following benchmarks: FollowBench~\citep{jiang2024followbench}, AdvancedIF~\citep{he2025advancedif}, and InfoBench~\citep{qin2024infobench}.
For evaluation, we use GPT-5-mini as the judge and report mean scores over five independent evaluation runs.
Detailed training and evaluation configurations and the training cost are provided in \aref{app:experimental_setup}.
\\
\subsection{Baselines}
We compare our method against baselines that differ in how they treat 
the seed prompts $x$ during training, grouped into three categories.

\paragraph{No prompt modification.} 
These methods train directly on the fixed WildChat prompts. Base rubrics use $\mathcal{R}(x)$ as the reward; WildChecklists~\citep{viswanathan2025checklists} uses high-quality rubrics generated offline; and policy-adaptive rubrics, motivated by DR Tulu~\citep{shao2025drtulu}, additionally inject criteria targeting the policy's current weaknesses at each iteration, while still leaving $x$ unchanged. As a non-RL reference, we also include Distillation \cite{hinton2015distilling, ho2023reasoning}, a supervised-fine-tuned variant of the policy trained on responses sampled from the Qwen3-8B tutor for the same prompts; this uses the tutor as a teacher rather than as an examiner-generator.

\paragraph{Policy-unaware prompt modification.} 
This category rewrites prompts to be harder independently of the policy. Evol-Instruct~\citep{xu2023wizardlm} applies predefined operations (e.g., adding constraints, deepening) to increase prompt complexity in a single offline pass, controlling for whether harder prompts alone improve learning, independent of policy-awareness.

\paragraph{Policy-adaptive prompt modification.}
This category adapts prompts based on the capabilities of the current policy.
EVA~\citep{ye2025evolving} selects prompts using the reward gap in scalar reward-model scores across policy responses and completely rewrites each selected prompt using Evol-Instruct templates. EVA differs from \emph{LLM-as-a-Tutor} in that it relies on pointwise scalar rewards rather than pairwise response judgments and rewrites the entire prompt rather than appending atomic constraints to the original prompt. For a fair comparison, we re-implement EVA within our rubric-based RL setup while preserving its prompt-selection criterion and evolution templates.


Further details for each baseline are provided in 
\aref{app:baseline_details}.

\subsection{Main results}\label{subsec:main_results}
\begin{table}[t]
\centering
\caption{Performance on instruction-following benchmarks under iterative setup. We use Qwen3-1.7B as a policy model and Qwen3-8B as the tutor and training-time judge, and GPT-5-mini judges the benchmarks. Scores are scaled to [0,100] and reported as mean±standard error of the mean over five independent evaluation runs. Best results in each column are highlighted in bold. \textbf{LLM-as-a-Tutor achieves the best average and outperforms adaptive and non-adaptive baselines on four of five metrics.}}
\vspace{4pt}

\label{tab:main_results}
\setlength{\tabcolsep}{2.5pt}
\begin{tabular}{lcccccc}
\toprule
& \multicolumn{2}{c}{\textbf{FollowBench}} & \multicolumn{2}{c}{\textbf{AdvancedIF}} & \textbf{InfoBench} & \multirow{2}{*}{\textbf{Average}} \\
\cmidrule(lr){2-3} \cmidrule(lr){4-5} \cmidrule(lr){6-6}
Method & HSR & SSR & Overall & Micro & DRFR & \\
\midrule
Qwen3-1.7B              & 35.34 \tiny{$\pm$0.519} & 58.43 \tiny{$\pm$0.253} & 13.08 \tiny{$\pm$0.410} & 66.48 \tiny{$\pm$0.290} & 72.28 \tiny{$\pm$0.760}          & 49.12 \tiny{$\pm$0.338} \\
\midrule
\multicolumn{7}{l}{\textit{No prompt modification}} \\
Distillation       & 34.67 \tiny{$\pm$0.502} & 57.15 \tiny{$\pm$0.435} & 11.59 \tiny{$\pm$0.240} & 58.27 \tiny{$\pm$0.420} & 68.58 \tiny{$\pm$0.450}             & 46.05 \tiny{$\pm$0.144}\\
Base rubrics            & 38.60 \tiny{$\pm$0.389} & 60.17 \tiny{$\pm$0.206} & 13.98 \tiny{$\pm$0.390} & 66.32 \tiny{$\pm$0.430} & 73.48 \tiny{$\pm$0.250}          & 50.51 \tiny{$\pm$0.265}\\
WildChecklists~\citep{viswanathan2025checklists} & 38.71 \tiny{$\pm$0.710} & 60.71 \tiny{$\pm$0.554} & 14.03 \tiny{$\pm$0.370} & 66.35 \tiny{$\pm$0.290} & 73.80 \tiny{$\pm$0.360} & 50.72 \tiny{$\pm$0.162}\\
Policy-adaptive rubrics & 39.63 \tiny{$\pm$0.595} & 61.30 \tiny{$\pm$0.455} & 13.93 \tiny{$\pm$0.540} & 66.15 \tiny{$\pm$0.320} & \textbf{74.19} \tiny{$\pm$0.530}          & 51.04 \tiny{$\pm$0.133}\\
\midrule
\multicolumn{7}{l}{\textit{Policy-unaware prompt modification}} \\
Evol-Instruct~\citep{xu2023wizardlm}           & 37.98 \tiny{$\pm$0.704} & 59.99 \tiny{$\pm$0.305} & 13.88 \tiny{$\pm$0.570} & 65.99 \tiny{$\pm$0.410} & 73.35 \tiny{$\pm$0.350}             & 50.24 \tiny{$\pm$0.267}\\
\midrule
\multicolumn{7}{l}{\textit{Policy-adaptive prompt modification}} \\
EVA~\citep{ye2025evolving}          & 39.14 \tiny{$\pm$0.918} & 61.04 \tiny{$\pm$0.622} & 14.93 \tiny{$\pm$0.500} & 66.66 \tiny{$\pm$0.180} & 73.43 \tiny{$\pm$0.290}          & 51.04 \tiny{$\pm$0.396}\\
\textbf{LLM-as-a-Tutor} & \textbf{40.91} \tiny{$\pm$0.547} & \textbf{62.28} \tiny{$\pm$0.527} & \textbf{15.07} \tiny{$\pm$0.370} & \textbf{67.97} \tiny{$\pm$0.120} & 73.59 \tiny{$\pm$0.100} & \textbf{51.96} \tiny{$\pm$0.226} \\
\bottomrule
\end{tabular}


\end{table}

\paragraph{LLM-as-a-Tutor outperforms both policy-unaware and prior policy-adaptive baselines.} 
As shown in \aref{tab:main_results}, LLM-as-a-Tutor achieves the highest average score across three instruction-following benchmarks and outperforms all baselines on four of five metrics.
The gain over the policy-unaware baseline confirms the value of adaptation itself, while the gain over prior policy-adaptive methods indicates that adapting at the level of \emph{prompts} provides a richer training signal that rubric-based RL alone cannot capture.
Under a DeepSeek-V4-Pro evaluation judge, LLM-as-a-Tutor again achieves the highest average (55.07 vs.\ 54.24 for Policy-adaptive rubrics) and outperforms all baselines on the same four of five metrics, which we detail in \aref{app:add_main}.


\paragraph{Policy-adaptive methods outperform policy-unaware ones.} 
Policy-unaware methods (Base rubrics, WildChecklists) yield modest improvements over the untrained policy, while methods that adapt to the policy at training time (Policy-adaptive rubrics, EVA) deliver larger gains and reach the same average (51.04) despite adapting to the rubric or prompt respectively. 
In contrast, Evol-Instruct, which rewrites prompts to be harder independently of the policy, underperforms even Base rubrics (50.24 vs.\ 50.51), suggesting that increased difficulty without policy-awareness can degrade training.

\paragraph{Prompt adaptation outperforms rubric adaptation.}
LLM-as-a-Tutor outperforms Policy-adaptive rubrics though both methods adapt to the current policy. 
The two methods differ in what they can change: rubrics determine how rollouts are scored, while prompts determine what rollouts the policy produces. 
When a prompt fails to elicit a quality gap among rollouts, no rubric, regardless of how well it is adapted, can produce a discriminative reward signal. 
In such cases, modifying the prompt itself can restore variance among rollouts.

\subsection{Ablation on policy-adaptive prompt selection}\label{subsec:ablation_select}
\paragraph{Policy-adaptive selection outperforms non-adaptive variants.}
\aref{tab:adapt_ablation} compares our method against non-adaptive variants that share the same append-based modification but differ in how target prompts are chosen. 
Applying constraints to all prompts (Always) or to a random 28\% (Random, matching the average constraint-added ratio across three epochs) both fall short of our adaptive variant, indicating that constraint addition is effective only when targeted at the right prompts. 
Adding constraints based on a larger Qwen3-8B model's responses instead of the policy's own (Mismatched) also underperforms, despite augmenting a larger fraction of prompts (47\% vs.\ our 28\%), showing that the relevant signal is the current policy's behavior, not difficulty alone. 

\paragraph{Pairwise judgments outperform rule-based selection.}
We compare our prompt selection using pairwise judgments with a variance-based alternative, following recent RLVR methods that use reward statistics as proxies for prompt learnability~\citep{bae2026online,jiang2025vcrl}.
Variance-based method selects prompts whose standard deviation of reward across the target policy's responses falls below a fixed threshold $\tau$ ($\tau=3$).
Similarity-based selection instead picks prompts whose rollouts have high embedding-based similarity.
Both alternatives underperform our method, suggesting that pairwise response judgments provide a more effective signal for identifying prompt saturation under continuous, rubric-based rewards.
We detail both alternatives in \aref{app:add_selection}. 
\vspace{5pt}



\begin{table}[t!]
\centering
\begin{minipage}[t]{0.58\textwidth}
\centering
\caption{Ablation on policy-adaptive prompt selection. All variants use the same append-based modification strategy but differ in how prompts are selected: Always (modify all prompts), Random (select prompts uniformly at random), Mismatched
(use responses from another model for constraint judgment), and two rule-based variants, Variance-based and Similarity-based (select prompts with the lowest reward-score standard deviation or the most similar rollouts). \textbf{Our method outperforms both policy-unaware variants and the rule-based selection on average.}\looseness=-1}
\vspace{4pt}
\label{tab:adapt_ablation}
\small
\setlength{\tabcolsep}{4pt}
\begin{tabular}{lcccccc}
\toprule
Method & Adaptive & FB & AdvIF & IB & Avg. \\
\midrule
Always & \xmark & 39.68 & 14.03 & 72.68 & 42.13 \\
Random & \xmark & 40.57 & 14.58 & 73.32 & 42.82 \\
Mismatched  & \tmark & 40.18 & 14.53 & \textbf{73.72} & 42.81 \\
Variance-based & \cmark & 40.01 & 13.98 & 73.62 & 42.54 \\
Similarity-based & \cmark & 40.58 & 13.88 & 73.36 & 42.61 \\
\textbf{Ours}  & \cmark & \textbf{40.91} & \textbf{15.07} & 73.59 & \textbf{43.19} \\
\bottomrule
\end{tabular}
\end{minipage}%
\hfill
\begin{minipage}[t]{0.4\textwidth}
\centering
\caption{Ablation on prompt modification strategies. We compare three strategies for modifying prompts: Reset (substitutes previous constraint between epochs), Rewrite (rewrites the entire prompt), and Append (Ours) (appends the constraint). \textbf{Append outperforms other modification variants (Reset and Rewrite) on all benchmarks.}
\vspace{3pt}
\looseness=-1}
\label{tab:method_ablation}
\small
\setlength{\tabcolsep}{1.5pt}
\begin{tabular}{lccccc}
\toprule
Method\,\,\,\,\,\, & FB & AdvIF & IB & Avg. \\
\midrule

Reset & 39.57 & 14.33 & 73.28 & 42.39 \\
Rewrite & 40.43 & 14.63 & 73.58 & 42.88 \\
\textbf{Append (Ours)} & \textbf{40.91} & \textbf{15.07} & \textbf{73.59} & \textbf{43.19} \\
\bottomrule
\end{tabular}
\end{minipage}
\end{table}
\begin{table}[t]
\centering
\caption{Performance of Qwen3-1.7B Base (without instruction tuning) as a policy model and Qwen3-8B as the tutor and training-time judge. Best results in each column are highlighted in bold. LLM-as-a-Tutor outperforms the baseline on all instruction-following benchmarks.}
\vspace{4pt}

\label{tab:base_result}
\setlength{\tabcolsep}{2.5pt}
\begin{tabular}{lcccccc}
\toprule
& \multicolumn{2}{c}{FollowBench} & \multicolumn{2}{c}{AdvancedIF} & InfoBench & \multirow{2}{*}{{Average}} \\
\cmidrule(lr){2-3} \cmidrule(lr){4-5} \cmidrule(lr){6-6}
Method & HSR & SSR & Overall & Micro & DRFR & \\
\midrule
Qwen3-1.7B Base  \,\,\,            & 8.99 & 20.79 & 2.14 & 26.84 & 35.45         & 18.84 \\
Base rubrics              & 17.86 & 39.29 & 6.02 & 52.89 & 65.86        & 36.38 \\
\textbf{Ours} & \textbf{18.76} & \textbf{41.72} & \textbf{6.57} & \textbf{54.05} & \textbf{67.07} & \textbf{37.63} \\
\bottomrule
\end{tabular}
\end{table}



\subsection{Ablation on prompt adaptation strategies}\label{subsec:ablation_rewrite}
\paragraph{Append outperforms other prompt modification methods.}
\aref{tab:method_ablation} compares three strategies for modifying a prompt once it is judged non-discriminative. Reset replaces the previous constraint with a new one each epoch, while Rewrite regenerates the entire prompt as a more challenging one. Neither guarantees monotonically increasing difficulty: discarding earlier constraints or rewriting from scratch can produce a prompt no harder than before. Rewrite has an additional drawback in that it drifts the prompt away from the original task distribution, weakening the alignment between the training signal and the instruction-following objective.
Append (Ours), in contrast, increases difficulty by adding a new constraint while retaining the original prompt and all previously added constraints.
Append achieves the best performance on all three benchmarks.

\subsection{Ablation on policy initialization}
\textbf{LLM-as-a-Tutor effectively aligns base models without prior instruction tuning.} We apply our framework directly to Qwen3-1.7B Base to test its robustness on a completely unaligned policy. As shown in \autoref{tab:base_result}, our method consistently outperforms the policy-unaware baseline across all benchmarks (37.63 vs. 36.38 on average), demonstrating that appending atomic constraints successfully reinstates discriminative reward signals even from scratch.

\section{Analysis}
\label{sec:analysis}
\begin{figure}[!t]
    \centering
    \small
    \begin{subfigure}[b]{0.32\textwidth}
        \centering
        \includegraphics[width=\linewidth]{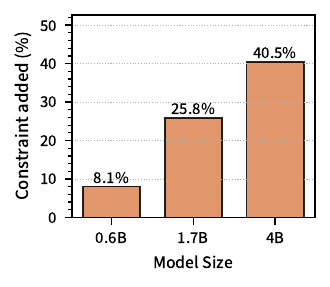}
        \subcaption{Constraint-added ratio across \
        different policy model sizes}
        \label{fig:constraint_density}
    \end{subfigure}%
    \hfill
    \begin{subfigure}[b]{0.32\textwidth}
        \centering
        \includegraphics[width=\linewidth]
        {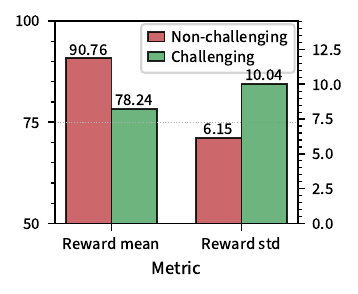}
        \subcaption{Reward statistics on challenging \ \quad and non-challenging prompts}
        \label{fig:difficulty_selection}
    \end{subfigure}
    \hfill
    \begin{subfigure}[b]{0.32\textwidth}
        \centering
        \includegraphics[width=\linewidth]{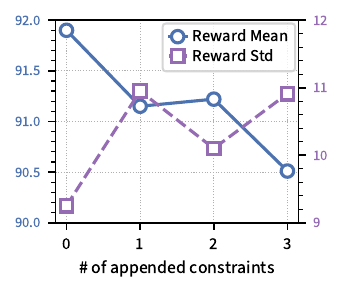}
        \subcaption{Base-rubric reward according to the number of appended constraints}
        \label{fig:cumulative_difficulty}
    \end{subfigure}
    \caption{\textbf{Analysis of the tutor's decisions and their effect on the reward signal. }
\textbf{(a)} Ratio of prompts identified as non-challenging by the tutor at its first pass (and thus augmented with constraints) across policies of different sizes. The ratio grows with policy size, showing that the tutor allocates more constraints to stronger policies. 
\textbf{(b)} Per-prompt reward mean and population standard deviation over eight rollouts, grouped by the tutor's first-pass challenging/non-challenging decision. Non-challenging prompts show higher mean and lower variance, confirming that the tutor's decision aligns with empirical difficulty. 
\textbf{(c)} Reward mean and standard deviation evaluated on the base rubrics only, as the number of appended constraints increases from 0 to 3. Prompts with appended constraints show a lower mean and a higher variance of base rubric scores than prompts without, indicating that constraints amplify the discriminative signal carried by the base rubrics.}
    \label{fig:analysis_panel}
\end{figure}

\subsection{Analysis of the tutor's behavior}\label{subsec:tutor_behavior}
\paragraph{Policy scaling drives denser constraint generation.}
\aref{fig:constraint_density} shows the ratio of prompts augmented with constraints at the tutor's first pass as the policy scales from 0.6B to 4B, with the tutor fixed at 8B.
The ratio grows monotonically with policy scale, from 8.1\% to 40.5\%. As the policy scales up, seed prompts become relatively easier and the policy produces high-quality rollouts with little variation among them, leading the tutor to judge more prompts as non-challenging.

\paragraph{Model-based decision correlates with empirical difficulty.}
\aref{fig:difficulty_selection} compares the per-prompt reward mean and standard deviation between prompts the tutor identifies as challenging and those it identifies as non-challenging.
Non-challenging prompts have a higher mean reward (90.76 vs.\ 78.24) and a lower mean per-prompt reward std (6.15 vs.\ 10.04), meaning the policy already produces high-quality and similar rollouts on them.
This separation shows that the tutor's decision effectively tracks the policy's empirical performance rather than being arbitrary.
Four or eight rollouts give a similar qualitative separation, while adaptation time scales linearly.
We therefore use two rollouts as a practical performance-cost trade-off rather than claiming that it is universally optimal, as detailed in \aref{app:add_tutor}.

\paragraph{Appended constraints amplify the discriminative power of base rubrics.}
\aref{fig:cumulative_difficulty} reports the per-prompt reward mean and standard deviation evaluated on the base rubrics only, as the number of appended constraints grows from 0 to 3.
Prompts with appended constraints show a lower base-rubric mean and a higher standard deviation than prompts without: the policy finds it harder to satisfy the same base criteria once additional constraints must be satisfied simultaneously, and the base rubrics continue to discriminate among rollouts rather than saturating as the policy improves.

\paragraph{Generated constraints are compatible and preserve intent.}
Two judges rate constraints from three tutors on compatibility, groundedness, self-verifiability, and intent preservation.
Compatibility exceeds 98\% and intent preservation 96\% for every tutor and judge, and the DeepSeek-V4-Pro tutor is uniformly higher across all dimensions.
We detail the setup in \aref{app:add_tutor}.

\subsection{Qualitative analysis}\label{subsec:qualitative}

\autoref{fig:qualitative_workplace_email} illustrates how a single atomic constraint converts a saturated prompt into one that \emph{induces} discriminative rollouts.
On the original prompt, all rollouts score $100$ on the base rubrics ($\sigma_{\text{base}} = 0$): the prompt is too easy for the current policy, every rollout succeeds, and the prompt no longer provides an informative reward signal.
Appending one clause that demands a \emph{specific named policy} (highlighted in \textcolor{red}{red}) breaks this saturation and splits the rollouts by a clear quality gap.
For example, Rollout A engages the new requirement directly, inserting an explicit \texttt{[Company Policy Name]} citation and scoring $100$; Rollout B retreats to a generic ``\emph{consult your supervisor}'' deflection without naming any policy and scores $60$.
We provide additional qualitative samples in \aref{app:additional_qualitative}.

\begin{figure}[t]
\centering
\small
\setlength{\tabcolsep}{8pt}
\renewcommand{\arraystretch}{1.25}
\begin{tabular}{p{0.45\textwidth} | p{0.45\textwidth}}
\toprule
\textbf{Before} ($x$) & \textbf{After} ($\tilde{x} = x \oplus c$) \\
\midrule
\textbf{Prompt:} Write a friendly workplace email to an employee: hope their daughter \texttt{Lila} is doing better, welcome her back at work, explain that company policy requires a \texttt{doctor's note} for absences over two days, and close with pleasantries.
&
\textbf{Prompt:} \textit{\{Original Prompt\}} + \emph{``Additionally, \textcolor{red}{include the specific name of the company policy} that requires a doctor's note for absences exceeding two days.''} \\
\midrule
$\mu_{\text{base}} = 100.0$, \quad $\sigma_{\text{base}} = 0.0$ & $\mu_{\text{base}} = 92.5$, \quad $\sigma_{\text{base}} = 14.9$ \\
\midrule
\multicolumn{2}{l}{\textbf{Sample rollouts on the adapted prompt} ($\tilde{x}$):} \\
\midrule
\multicolumn{2}{l}{\parbox{0.94\textwidth}{\vspace{2pt}
\textcolor{green!60!black}{\ding{52}} \textbf{Rollout A} ($s_{\text{base}}=100.0$):

Subject: Update on Lila's Recovery \& Policy Reminder

Hi [Employee's Name],
[...]
As per company policy, we will require a doctor's note prior to approving sick time requests for absences exceeding two days. \textcolor{red}{This is in line with [Company Policy Name], which outlines the requirement for medical documentation to ensure proper care and prevent unnecessary absences.}

[...]
\vspace{2pt}}} \\
\midrule
\multicolumn{2}{l}{\parbox{0.94\textwidth}{\vspace{2pt}
\textcolor{red}{\ding{56}} \textbf{Rollout B} ($s_{\text{base}}=60.0$):

Subject: Hope Your Daughter, Lila, Is Doing Well!

Hi [Employee's Name],

[...]
\textcolor{red}{As per company policy}, we require a doctor's note prior to approving sick time requests for absences exceeding two days. \textcolor{red}{Please ensure you consult with your supervisor to obtain the necessary documentation,} and feel free to reach out if you need assistance.

[...]
\vspace{2pt}}} \\
\bottomrule
\end{tabular}
\caption{%
\textbf{A single atomic constraint reinstates discriminative reward signal on a non-challenging prompt.}
Before adding a constraint $c$, all rollouts score $100$ (std $=0$); after, reward mean drops to $92.5$ with std $14.9$.
Rollout A explicitly cites a named policy (satisfying $c$); Rollout B deflects to ``consult your supervisor'' without naming one.
}
\label{fig:qualitative_workplace_email}
\end{figure}

\section{Discussion}
\label{sec:discussion}

\paragraph{LLM-as-a-Tutor as a potentially stronger form of distillation.}
Rubric-based RL succeeds because fine-grained reward signals nudge the policy toward responses it is already capable of producing but does not by default~\citep{gunjal2026rubrics, viswanathan2025checklists}.
LLM-as-a-Tutor extends this principle to the prompt distribution, using the tutor's discriminative capability to identify where the policy requires harder prompts and to construct them.
We believe this can be viewed as a form of distillation that transfers adaptive feedback rather than target responses.
Unlike SFT distillation, which is bounded by the teacher's outputs~\citep{hinton2015distilling, ho2023reasoning, magister2023teaching}, our tutor produces no target response and instead shapes the training signal so the student discovers the capability through RL.
The student's ceiling is therefore not the tutor's ability to answer the prompt, but only its ability to discriminate between two answers, a regime in which LLMs are consistently sharper~\citep{liu2024aligning, liusie2024comparative, zheng2023judging}.

\paragraph{Generalizing the LLM-as-a-Tutor framework beyond constraint-additive tasks.}
Our choice to append atomic constraints aligns with how complex instruction following is operationalized in benchmarks such as FollowBench~\citep{jiang2024followbench}, where difficulty is defined by the number of constraints a response must simultaneously satisfy.
This alignment is by design, since constraint count is the natural notion of difficulty for non-verifiable instruction following.
Our evaluation targets constraint-heavy instruction following, and our FollowBench subset alone spans 13 task families.
The gains also transfer to Creative Writing v3 instruction adherence (5.38 vs.\ 5.21 for Base rubrics), as detailed in \aref{app:add_generalize}.
Other domains may admit different modification templates: added reasoning steps for multi-step reasoning~\citep{guo2025deepseekr1}, additional edge cases for code, tighter sourcing requirements for factual QA~\citep{shao2025drtulu}, which we leave to future work.

\paragraph{Prompt growth and constraint removal.}
Over three epochs, the tutor appends 0.83 constraints per prompt on average, and only 6\% of prompts receive one in every epoch.
The tutor acts as the stopping mechanism, adding a constraint only when the policy has re-saturated the current prompt and saying no when in doubt.
Because each constraint carries its own independently scored rubric criteria, a constraint whose criteria show zero score variance across rollouts can be removed or merged, as DR Tulu does for zero-variance rubrics~\citep{shao2025drtulu}, which we leave to future work.
We report the full distribution of appended constraints in \aref{app:add_growth}.
\section{Conclusion}
\label{sec:conclusion}




We introduce \textit{LLM-as-a-Tutor}, which addresses the policy--prompt mismatch in rubric-based non-verifiable RL by reusing the judge LLM to also adapt the training prompts.
When seed prompts fail to elicit quality variance among rollouts, the tutor appends targeted constraints, yielding a self-calibrating curriculum that tracks the policy's capability without external difficulty schedules.
Across three instruction-following benchmarks, our method outperforms both policy-unaware and prior policy-adaptive baselines on average, suggesting prompt adaptation as an underexplored axis of policy-awareness in non-verifiable RL, complementary to rubric adaptation.
We believe this framework can extend to other domains with an appendable notion of difficulty, such as reasoning steps, edge cases, or sourcing requirements, which we leave to future work.

\clearpage
\begin{ack}
This work was supported by grants funded by the Korean government (Ministry of Science and ICT, MSIT) through the Institute of Information \& Communications Technology Planning \& Evaluation (IITP) ([No. RS-2024-00457882, National AI Research Lab Project, 10\%], [No. RS-2026-25522672, Development of Unified Reasoning Technology Mimicking Human Cognition for Hierarchical Understanding and Unbounded Problem Solving, 30\%]) and the National IT Industry Promotion Agency (NIPA) ([No. RS-2026-25621604, 30\%]). This work was also conducted as part of the Sovereign AI Foundation Model Project (GPU Track), organized by MSIT and supported by NIPA, Republic of Korea ([PJT-26-010017, 30\%]).
The authors declare no competing interests.
\end{ack}

\clearpage
\bibliographystyle{plainnat}
\bibliography{reference}

@inproceedings{gunjal2026rubrics,
  title     = {Rubrics as Rewards: Reinforcement Learning Beyond Verifiable Domains},
  author    = {Gunjal, Anisha and Wang, Anthony and Lau, Elaine and Nath, Vaskar and He, Yunzhong and Liu, Bing and Hendryx, Sean M.},
  booktitle = {The Fourteenth International Conference on Learning Representations},
  year      = {2026},
  url       = {https://openreview.net/forum?id=c1bTcrDmt4}
}

@inproceedings{viswanathan2025checklists,
  title     = {Checklists Are Better Than Reward Models For Aligning Language Models},
  author    = {Viswanathan, Vijay and Sun, Yanchao and Kong, Xiang and Cao, Meng and Neubig, Graham and Wu, Tongshuang},
  booktitle = {Advances in Neural Information Processing Systems},
  year      = {2025},
  note      = {Spotlight},
  url       = {https://openreview.net/forum?id=RPRqKhjrr6}
}

@article{shao2025drtulu,
  title   = {DR Tulu: Reinforcement Learning with Evolving Rubrics for Deep Research},
  author  = {Shao, Rulin and Asai, Akari and Shen, Shannon Zejiang and Ivison, Hamish and Kishore, Varsha and Zhuo, Jingming and Zhao, Xinran and Park, Molly and Finlayson, Samuel G. and Sontag, David and Murray, Tyler and Min, Sewon and Dasigi, Pradeep and Soldaini, Luca and Brahman, Faeze and Yih, Wen-tau and Wu, Tongshuang and Zettlemoyer, Luke and Kim, Yoon and Hajishirzi, Hannaneh and Koh, Pang Wei},
  journal = {arXiv preprint arXiv:2511.19399},
  year    = {2025}
}

@inproceedings{ye2025evolving,
  title     = {Scalable Reinforcement Post-Training Beyond Static Human Prompts: Evolving Alignment via Asymmetric Self-Play},
  author    = {Ye, Ziyu and Agarwal, Rishabh and Liu, Tianqi and Joshi, Rishabh and Velury, Sarmishta and Le, Quoc V. and Tan, Qijun and Liu, Yuan},
  booktitle = {Proceedings of the 42nd International Conference on Machine Learning},
  year      = {2025},
  url       = {https://arxiv.org/abs/2411.00062}
}

@article{kuba2025language,
  title   = {Language Self-Play For Data-Free Training},
  author  = {Kuba, Jakub Grudzien and Gu, Mengting and Ma, Qi and Tian, Yuandong and Mohan, Vijai and Chen, Jason},
  journal = {arXiv preprint arXiv:2509.07414},
  year    = {2025}
}

@inproceedings{zhao2025absolute,
  title     = {Absolute Zero: Reinforced Self-Play Reasoning with Zero Data},
  author    = {Zhao, Andrew and Wu, Yiran and Yue, Yang and Wu, Tong and Xu, Quentin and Yue, Yang and Lin, Matthieu and Wang, Shenzhi and Wu, Qingyun and Zheng, Zilong and Huang, Gao},
  booktitle = {Advances in Neural Information Processing Systems},
  year      = {2025},
  note      = {Spotlight},
  url       = {https://arxiv.org/abs/2505.03335}
}

@article{huang2025rzero,
  title   = {R-Zero: Self-Evolving Reasoning {LLM} from Zero Data},
  author  = {Huang, Chengsong and Yu, Wenhao and Wang, Xiaoyang and Zhang, Hongming and Li, Zongxia and Li, Ruosen and Huang, Jiaxin and Mi, Haitao and Yu, Dong},
  journal = {arXiv preprint arXiv:2508.05004},
  year    = {2025}
}

@article{liu2025spice,
  title   = {{SPICE}: Self-Play In Corpus Environments Improves Reasoning},
  author  = {Liu, Bo and Jin, Chuanyang and Kim, Seungone and Yuan, Weizhe and Zhao, Wenting and Kulikov, Ilia and Li, Xian and Sukhbaatar, Sainbayar and Lanchantin, Jack and Weston, Jason},
  journal = {arXiv preprint arXiv:2510.24684},
  year    = {2025}
}

@inproceedings{kim2024prometheus,
  title     = {Prometheus: Inducing Fine-Grained Evaluation Capability in Language Models},
  author    = {Seungone Kim and Jamin Shin and Yejin Cho and Joel Jang and Shayne Longpre and Hwaran Lee and Sangdoo Yun and Seongjin Shin and Sungdong Kim and James Thorne and Minjoon Seo},
  booktitle = {The Twelfth International Conference on Learning Representations (ICLR)},
  year      = {2024},
  url       = {https://openreview.net/forum?id=8euJaTveKw}
}

@inproceedings{zheng2023judging,
  title     = {Judging {LLM}-as-a-Judge with {MT}-Bench and Chatbot Arena},
  author    = {Lianmin Zheng and Wei-Lin Chiang and Ying Sheng and Siyuan Zhuang and Zhanghao Wu and Yonghao Zhuang and Zi Lin and Zhuohan Li and Dacheng Li and Eric P. Xing and Hao Zhang and Joseph E. Gonzalez and Ion Stoica},
  booktitle = {Advances in Neural Information Processing Systems 36 (NeurIPS 2023) Datasets and Benchmarks Track},
  year      = {2023},
  url       = {https://proceedings.neurips.cc/paper_files/paper/2023/hash/91f18a1287b398d378ef22505bf41832-Abstract-Datasets_and_Benchmarks.html}
}

@inproceedings{jiang2024followbench,
  title={Followbench: A multi-level fine-grained constraints following benchmark for large language models},
  author={Jiang, Yuxin and Wang, Yufei and Zeng, Xingshan and Zhong, Wanjun and Li, Liangyou and Mi, Fei and Shang, Lifeng and Jiang, Xin and Liu, Qun and Wang, Wei},
  booktitle={Proceedings of the 62nd Annual Meeting of the Association for Computational Linguistics (Volume 1: Long Papers)},
  pages={4667--4688},
  year={2024}
}

@inproceedings{qin2024infobench,
  title={Infobench: Evaluating instruction following ability in large language models},
  author={Qin, Yiwei and Song, Kaiqiang and Hu, Yebowen and Yao, Wenlin and Cho, Sangwoo and Wang, Xiaoyang and Wu, Xuansheng and Liu, Fei and Liu, Pengfei and Yu, Dong},
  booktitle={Findings of the Association for Computational Linguistics: ACL 2024},
  pages={13025--13048},
  year={2024}
}

@article{he2025advancedif,
  title={Advancedif: Rubric-based benchmarking and reinforcement learning for advancing llm instruction following},
  author={He, Yun and Li, Wenzhe and Zhang, Hejia and Li, Songlin and Mandyam, Karishma and Khosla, Sopan and Xiong, Yuanhao and Wang, Nanshu and Peng, Xiaoliang and Li, Beibin and others},
  journal={arXiv preprint arXiv:2511.10507},
  year={2025}
}

@inproceedings{zhao2024wildchat,
  title     = {WildChat: 1{M} ChatGPT Interaction Logs in the Wild},
  author    = {Zhao, Wenting and Ren, Xiang and Hessel, Jack and Cardie, Claire and Choi, Yejin and Deng, Yuntian},
  booktitle = {The Twelfth International Conference on Learning Representations (ICLR)},
  year      = {2024},
  url       = {https://openreview.net/forum?id=Bl8u7ZRlbM}
}

@article{shao2024deepseekmath,
  title={Deepseekmath: Pushing the limits of mathematical reasoning in open language models},
  author={Shao, Zhihong and Wang, Peiyi and Zhu, Qihao and Xu, Runxin and Song, Junxiao and Bi, Xiao and Zhang, Haowei and Zhang, Mingchuan and Li, YK and Wu, Yang and others},
  journal={arXiv preprint arXiv:2402.03300},
  year={2024}
}

@article{yang2025qwen3,
  title={Qwen3 Technical Report},
  author={Yang, An and Li, Anfeng and Yang, Baosong and Zhang, Beichen and Hui, Binyuan and Zheng, Bo and Yu, Bowen and Gao, Chang and Huang, Chengen and Lv, Chenxu and others},
  journal={arXiv preprint arXiv:2505.09388},
  year={2025}
}

@article{casper2023open,
  title   = {Open Problems and Fundamental Limitations of Reinforcement Learning from Human Feedback},
  author  = {Casper, Stephen and Davies, Xander and Shi, Claudia and Gilbert, Thomas Krendl and Scheurer, J{\'e}r{\'e}my and Rando, Javier and Freedman, Rachel and Korbak, Tomasz and Lindner, David and Freire, Pedro and Wang, Tony and Marks, Samuel and S{\'e}gerie, Charbel-Rapha{\"e}l and Carroll, Micah and Peng, Andi and Christoffersen, Phillip and Damani, Mehul and Slocum, Stewart and Anwar, Usman and Siththaranjan, Anand and Nadeau, Max and Michaud, Eric J. and Pfau, Jacob and Krasheninnikov, Dmitrii and Chen, Xin and Langosco, Lauro and Hase, Peter and B{\i}y{\i}k, Erdem and Dragan, Anca and Krueger, David and Sadigh, Dorsa and Hadfield-Menell, Dylan},
  journal = {Transactions on Machine Learning Research},
  year    = {2023},
  url     = {https://openreview.net/forum?id=bx24KpJ4Eb}
}

@article{xu2023wizardlm,
  title={WizardLM: Empowering large pre-trained language models to follow complex instructions},
  author={Xu, Can and Sun, Qingfeng and Zheng, Kai and Geng, Xiubo and Zhao, Pu and Feng, Jiazhan and Tao, Chongyang and Lin, Qingwei and Jiang, Daxin},
  journal={arXiv preprint arXiv:2304.12244},
  year={2023}
}

@inproceedings{christiano2017deep,
  author    = {Christiano, Paul F. and Leike, Jan and Brown, Tom and Martic, Miljan and Legg, Shane and Amodei, Dario},
  title     = {Deep Reinforcement Learning from Human Preferences},
  booktitle = {Advances in Neural Information Processing Systems 30 (NIPS 2017)},
  editor    = {Guyon, I. and Von Luxburg, U. and Bengio, S. and Wallach, H. and Fergus, R. and Vishwanathan, S. and Garnett, R.},
  pages     = {4299--4307},
  year      = {2017},
  publisher = {Curran Associates, Inc.},
  url       = {https://proceedings.neurips.cc/paper/2017/hash/d5e2c0adad503c91f91df240d0cd4e49-Abstract.html}
}

@inproceedings{ouyang2022training,
  author    = {Ouyang, Long and Wu, Jeffrey and Jiang, Xu and Almeida, Diogo and Wainwright, Carroll L. and Mishkin, Pamela and Zhang, Chong and Agarwal, Sandhini and Slama, Katarina and Ray, Alex and Schulman, John and Hilton, Jacob and Kelton, Fraser and Miller, Luke and Simens, Maddie and Askell, Amanda and Welinder, Peter and Christiano, Paul F. and Leike, Jan and Lowe, Ryan},
  title     = {Training Language Models to Follow Instructions with Human Feedback},
  booktitle = {Advances in Neural Information Processing Systems 35 (NeurIPS 2022)},
  editor    = {Koyejo, S. and Mohamed, S. and Agarwal, A. and Belgrave, D. and Cho, K. and Oh, A.},
  pages     = {27730--27744},
  year      = {2022},
  publisher = {Curran Associates, Inc.},
  url       = {https://proceedings.neurips.cc/paper_files/paper/2022/hash/b1efde53be364a73914f58805a001731-Abstract-Conference.html}
}

@article{tzannetos2023proximal,
  title={Proximal Curriculum for Reinforcement Learning Agents},
  author={Tzannetos, Georgios and Gomes Ribeiro, B{\'a}rbara and Kamalaruban, Parameswaran and Singla, Adish},
  journal={Transactions on Machine Learning Research},
  year={2023}
}

@article{guo2025deepseekr1,
  title={Deepseek-r1: Incentivizing reasoning capability in llms via reinforcement learning},
  author={Guo, Daya and Yang, Dejian and Zhang, Haowei and Song, Junxiao and Wang, Peiyi and Zhu, Qihao and Xu, Runxin and Zhang, Ruoyu and Ma, Shirong and Bi, Xiao and others},
  journal={arXiv preprint arXiv:2501.12948},
  year={2025}
}

@inproceedings{liu2024aligning,
  title={Aligning with Human Judgement: The Role of Pairwise Preference in Large Language Model Ev  aluators},
  author={Liu, Yinhong and Zhou, Han and Guo, Zhijiang and Shareghi, Ehsan and Vuli{\'c}, Ivan and Korhonen, Anna and Collier, Nigel},
  booktitle={First Conference on Language Modeling (COLM)},
  year={2024}
}

@inproceedings{liusie2024comparative,
  title={{LLM} Comparative Assessment: Zero-shot {NLG} Evaluation through Pairwise Comparisons using Large Language Models},
  author={Liusie, Adian and Manakul, Potsawee and Gales, Mark},
  booktitle={Proceedings of the 18th Conference of the European Chapter of the Association for Computational Linguistics (EACL)},
  pages={139--151},
  year={2024}
}

@inproceedings{bae2026online,
  title={Online difficulty filtering for reasoning oriented reinforcement learning},
  author={Bae, Sanghwan and Hong, Jiwoo and Lee, Min Young and Kim, Hanbyul and Nam, JeongYeon and Kwak, Donghyun},
  booktitle={Proceedings of the 19th Conference of the European Chapter of the Association for Computational Linguistics (Volume 1: Long Papers)},
  pages={700--719},
  year={2026},
  url={https://aclanthology.org/2026.eacl-long.30/}
}

@article{jiang2025vcrl,
  title={Vcrl: Variance-based curriculum reinforcement learning for large language models},
  author={Jiang, Guochao and Feng, Wenfeng and Quan, Guofeng and Hao, Chuzhan and Zhang, Yuewei and Liu, Guohua and Wang, Hao},
  journal={arXiv preprint arXiv:2509.19803},
  year={2025}
}

@article{liu2025openrubrics,
  title={Openrubrics: Towards scalable synthetic rubric generation for reward modeling and llm alignment},
  author={Liu, Tianci and Xu, Ran and Yu, Tony and Hong, Ilgee and Yang, Carl and Zhao, Tuo and Wang, Haoyu},
  journal={arXiv preprint arXiv:2510.07743},
  year={2025}
}

@inproceedings{sheng2025hybridflow,
  title={Hybridflow: A flexible and efficient rlhf framework},
  author={Sheng, Guangming and Zhang, Chi and Ye, Zilingfeng and Wu, Xibin and Zhang, Wang and Zhang, Ru and Peng, Yanghua and Lin, Haibin and Wu, Chuan},
  booktitle={Proceedings of the Twentieth European Conference on Computer Systems},
  pages={1279--1297},
  year={2025}
}

@article{rafailov2023direct,
  title={Direct preference optimization: Your language model is secretly a reward model},
  author={Rafailov, Rafael and Sharma, Archit and Mitchell, Eric and Manning, Christopher D and Ermon, Stefano and Finn, Chelsea},
  journal={Advances in neural information processing systems},
  volume={36},
  pages={53728--53741},
  year={2023}
}

@inproceedings{ho2023reasoning,
  title     = {Large Language Models Are Reasoning Teachers},
  author    = {Ho, Namgyu and Schmid, Laura and Yun, Se-Young},
  booktitle = {Proceedings of the 61st Annual Meeting of the Association for Computational Linguistics (Volume 1: Long Papers)},
  pages     = {14852--14882},
  year      = {2023}
}

@inproceedings{magister2023teaching,
  title     = {Teaching Small Language Models to Reason},
  author    = {Magister, Lucie Charlotte and Mallinson, Jonathan and Adamek, Jakub and Malmi, Eric and Severyn, Aliaksei},
  booktitle = {Proceedings of the 61st Annual Meeting of the Association for Computational Linguistics (Volume 2: Short Papers)},
  pages     = {1773--1781},
  year      = {2023}
}

@article{hinton2015distilling,
  title   = {Distilling the Knowledge in a Neural Network},
  author  = {Hinton, Geoffrey and Vinyals, Oriol and Dean, Jeff},
  journal = {arXiv preprint arXiv:1503.02531},
  year    = {2015}
}

@book{sutton2018rl,
  title     = {Reinforcement Learning: An Introduction},
  author    = {Sutton, Richard S. and Barto, Andrew G.},
  year      = {2018},
  edition   = {Second},
  publisher = {The MIT Press},
  address   = {Cambridge, MA}
}

@article{williams1992reinforce,
  title   = {Simple Statistical Gradient-Following Algorithms for Connectionist Reinforcement Learning},
  author  = {Williams, Ronald J.},
  journal = {Machine Learning},
  volume  = {8},
  pages   = {229--256},
  year    = {1992},
  publisher = {Springer}
}

@article{schulman2017ppo,
  title         = {Proximal Policy Optimization Algorithms},
  author        = {Schulman, John and Wolski, Filip and Dhariwal, Prafulla and Radford, Alec and Klimov, Oleg},
  journal       = {arXiv preprint arXiv:1707.06347},
  year          = {2017}
}

@article{stiennon2020learning,
  title={Learning to summarize with human feedback},
  author={Stiennon, Nisan and Ouyang, Long and Wu, Jeffrey and Ziegler, Daniel and Lowe, Ryan and Voss, Chelsea and Radford, Alec and Amodei, Dario and Christiano, Paul F},
  journal={Advances in neural information processing systems},
  volume={33},
  pages={3008--3021},
  year={2020}
}

@inproceedings{liu2023geval,
  title     = {{G-Eval}: {NLG} Evaluation using {GPT-4} with Better Human Alignment},
  author    = {Liu, Yang and Iter, Dan and Xu, Yichong and Wang, Shuohang and Xu, Ruochen and Zhu, Chenguang},
  booktitle = {Proceedings of the 2023 Conference on Empirical Methods in Natural Language Processing (EMNLP)},
  pages     = {2511--2522},
  year      = {2023}
}

@inproceedings{bengio2009curriculum,
  title     = {Curriculum Learning},
  author    = {Bengio, Yoshua and Louradour, J{\'e}r{\^o}me and Collobert, Ronan and Weston, Jason},
  booktitle = {Proceedings of the 26th Annual International Conference on Machine Learning (ICML)},
  pages     = {41--48},
  year      = {2009}
}

@inproceedings{kwon2023efficient,
  title     = {Efficient Memory Management for Large Language Model Serving with {PagedAttention}},
  author    = {Kwon, Woosuk and Li, Zhuohan and Zhuang, Siyuan and Sheng, Ying and Zheng, Lianmin and Yu, Cody Hao and Gonzalez, Joseph E. and Zhang, Hao and Stoica, Ion},
  booktitle = {Proceedings of the ACM SIGOPS 29th Symposium on Operating Systems Principles (SOSP)},
  year      = {2023}
}
\clearpage
\appendix

\section{Limitations}
\label{app:limitations}

\paragraph{Scope of evaluated policy--tutor configurations.}
Our main results in \aref{tab:main_results} are reported for a single policy--tutor pair (Qwen3-1.7B policy with Qwen3-8B tutor), and we do not run the full benchmark suite across additional model families or larger policy sizes due to compute constraints. The mechanism behind LLM-as-a-Tutor, however, is not specific to this pair: \aref{fig:constraint_density} sweeps the policy from 0.6B to 4B with the tutor held fixed and shows that the constraint-addition ratio rises monotonically with policy capability (8.1\% $\rightarrow$ 25.8\% $\rightarrow$ 40.5\%). The adaptation procedure thus tracks policy capability rather than locking onto a particular checkpoint, suggesting the method should generalize to other policies whose capability falls within the tutor's discriminative range.

\paragraph{Dependence on tutor capability.}
The method assumes the tutor can perform pairwise discrimination and atomic constraint generation reliably; a weaker tutor produces noisier saturation judgments and less targeted constraints, as the Mismatched baseline in \aref{tab:adapt_ablation} illustrates when judgments are made on a stronger model's rollouts rather than the policy's.
Because pairwise judgment and constraint generation are both reasoning-heavy subtasks, improvements in tutor reasoning capability translate directly into stronger adaptation, and the same pipeline should yield correspondingly stronger results as more capable open models become available.

\paragraph{Additional inference cost from tutor invocations.}
LLM-as-a-Tutor introduces additional tutor calls during training.
Measured end to end, training takes 19.22\,h with the tutor versus 16.73\,h for Base rubrics, a 15\% wall-clock training time overhead that we break down in \aref{app:compute}.
Per training prompt and epoch, our method requires two sample rollouts plus a single tutor call that makes the pairwise saturation judgment and, when the prompt is judged saturated, also generates the constraint and its rubric criteria.
For comparison, the existing rubric-based RL pipeline already consumes 8 policy rollouts and $8 \times N$ judge invocations per prompt, where $N$ is the number of rubric criteria (typically 5--10).
The tutor therefore adds 2 rollouts and 1 call per prompt per epoch, 3 to 6\% of the pipeline's LLM invocations before counting judge calls on appended criteria.
Each appended criterion is scored on all 8 rollouts at every later training step, which accounts for further judge calls.

\paragraph{Scope of task families.}
We evaluate only constraint-heavy instruction following, where difficulty is naturally instantiated as the number of requirements a response must satisfy.
Appending constraints may help less for tasks such as factual QA, multi-turn dialogue, or safety, which we leave to future work.

\paragraph{Prompts that are too hard for append-only adaptation.}
Append-only adaptation raises difficulty and never lowers it, so it cannot help a prompt that is already too hard for the policy.
Rubric rewards provide a dense, graded reward signal, so even extremely difficult prompts typically yield partial-credit signals rather than uniformly zero rewards.
Among 3,891 evaluable prompts from our WildChat seed set, scored on eight rollouts under the frozen initial policy, only 2.34\% have zero reward variance (\aref{tab:reward_std_dist}).
We exclude the 103 of 3,994 prompts with truncated or unparsable rollouts.
Of these 91 prompts, 84 receive a score of 100 on every rollout, 7 receive another constant score, and none receive zero reward on every rollout.

\begin{table}[h]
\centering
\small
\caption{Distribution of per-prompt reward std on the original prompts under the frozen policy.}
\label{tab:reward_std_dist}
\begin{tabular}{lrrr}
\toprule
Per-prompt reward std & Count & Proportion & Cumulative \\
\midrule
$=0$ & 91 & 2.34\% & 2.34\% \\
$(0, 1)$ & 204 & 5.24\% & 7.58\% \\
$[1, 3)$ & 671 & 17.24\% & 24.83\% \\
$[3, 5)$ & 526 & 13.52\% & 38.34\% \\
$[5, 10)$ & 1,245 & 32.00\% & 70.34\% \\
$\geq 10$ & 1,154 & 29.66\% & 100.00\% \\
\midrule
Total & 3,891 & 100.00\% & \\
\bottomrule
\end{tabular}
\end{table}

\section{Broader impacts}
\label{app:broader_impacts}

Our work improves the instruction-following capability of language models through policy-adaptive prompt modification in non-verifiable RL. The positive impact is more reliable instruction following: models that better track user-specified constraints are more useful for downstream applications such as writing assistance, document drafting, and other open-ended tasks where rubric-based quality matters. The training-time mechanism is also general; it could be combined with safety- or helpfulness-oriented rubrics to improve adherence to those objectives.

The same mechanism, however, has dual-use potential. An adaptation procedure that elicits stronger compliance with arbitrary instructions can equally elicit stronger compliance with harmful ones if the seed corpus or rubrics target undesirable objectives.
Because LLM-as-a-Tutor does not encode any notion of what the appended constraints should optimize for, it inherits whatever values are implicit in the seed prompts and the tutor's judgments.
We mitigate this in our experiments by using a public, non-adversarial prompt corpus (WildChat) and rubrics aimed at general response quality, but practitioners applying our method should ensure both the seed distribution and the tutor model are aligned with their intended deployment context. We do not release model checkpoints trained with this method, which limits direct misuse risk from this work.

\section{Pseudocode}\label{app:pseudo_code}
\begin{algorithm}[h]
\caption{Overall training pipeline}
\label{alg:tutor}
\begin{algorithmic}[1]
\Require Seed prompts $\mathcal{P} = \{x_n\}_{n=1}^N$, Policy model $\pi_\theta$, Tutor and Reward model $\pi_\phi$, Training epochs $T$, Rollouts per prompt $G$
\State $R(x_n) \gets \pi_\phi(\cdot \mid x_n)$ \textbf{for all} $n$ \Comment{Generate base rubric for each prompt}
\State $\mathcal{D} \gets \{(x_n,\, R(x_n))\}_{n=1}^N$
\For{$t = 1$ \textbf{to} $T$}
    \For{$(x, R(x)) \in \mathcal{D}$}
        \State $y^{(1)}, y^{(2)} \sim \pi_\theta(\cdot \mid x)$ \Comment{Sample policy rollouts}
        \State $(\textit{decision},\, c,\, R_c) \gets \pi_\phi(\cdot \mid x,y^{(1)},y^{(2)})$ \Comment{Judge and adapt prompt}

        \If{$\textit{decision} = yes$}
            \State $\tilde{x} \gets x \oplus c$, \quad $\tilde{R}(x)$ $\gets R(x) \cup R_c$ \Comment{Append constraint and rubric}
            \State Update $\mathcal{D}$ with $(\tilde{x}, \tilde{R}(x))$
        \EndIf
    \EndFor
    \\
    \For{minibatch $\mathcal{B} \subset \mathcal{D}$} \Comment{Train policy via rubric-based RL}
        \For{$(x, R(x)) \in \mathcal{B}$}
            \State $\{y^{(i)}\}_{i=1}^G \sim \pi_\theta(\cdot \mid x)$ 
            \State $s_j^{(i)} \gets \pi_\phi(\cdot \mid x,y^{(i)},r_j)$ \textbf{for} $i=1,\dots,G$, $(r_j,w_j) \in R(x)$
            \State $s^{(i)} \gets \sum_{(r_j,w_j)\in R(x)} w_j\,s_j^{(i)}$ \textbf{for} $i=1,\dots,G$
        \EndFor
        \State Update $\pi_\theta$ via GRPO using $\{s^{(i)}\}_{i=1}^G$
    \EndFor
\EndFor
\State \Return $\pi_\theta$
\end{algorithmic}
\end{algorithm}

\section{Experimental setup}\label{app:experimental_setup}

\subsection{Training}\label{app:training}
We implement all experiments using VeRL~\citep{sheng2025hybridflow} framework.
Of the 4,000 sampled WildChat prompts, 3,994 remain after filtering overlong prompts.
Training hyperparameters are summarized in \aref{tab:train_hparams}.

\begin{table}[h]
\centering
\small
\setlength{\tabcolsep}{6pt}
\captionsetup{skip=10pt}
\begin{tabular}{l l}
\toprule
Hyperparameter & Value \\
\midrule
Algorithm                & GRPO \\

Learning rate            & 5e-6 \\
Learning rate schedule            & constant \\
LR warmup steps          & 25 \\

Train batch size         & 32 \\
PPO mini-batch size      & 8 (4-step off-policy) \\

Rollouts per prompt      & 8 \\
Epochs & 3 (Iterative) \\

Clip ratio               & (0.2, 0.2) \\
KL loss coefficient      & 0.001 \\

Max prompt length        & 8192 \\
Max response length      & 4096 \\
Filter overlong prompts  & true \\
Temperature              & 0.6 \\
Top-$p$                  & 0.95 \\
Top-$k$                  & 20 \\
\bottomrule
\end{tabular}
\caption{Training hyperparameters.}
\label{tab:train_hparams}
\end{table}

\subsection{Evaluation}\label{app:evaluation}
We evaluate all policy models on three instruction-following benchmarks (FollowBench, AdvancedIF, InfoBench) using a unified vLLM-based generation pipeline~\cite{kwon2023efficient}.
Because LLM-based judges introduce non-trivial scoring stochasticity, we report the mean over $5$ independent evaluation runs for every benchmark.
For FollowBench, we evaluate on the \emph{content}, \emph{situation}, \emph{style}, \emph{format}, and \emph{mixed} categories, excluding the \emph{example} category.
For AdvancedIF, we restrict evaluation to the \emph{single-turn complex instruction-following} split. Otherwise, we follow the standard evaluation protocol for each benchmark, including the metric calculation.
Sampling parameters and the judge model used for each benchmark are summarized in \aref{tab:eval_settings}. 
We additionally re-evaluate the main results with DeepSeek-V4-Pro as the judge in \aref{app:add_main}.

\begin{table}[h]
\centering
\small
\setlength{\tabcolsep}{6pt}
\captionsetup{skip=10pt}
\begin{tabular}{l l}
\toprule
Hyperparameter & Value \\
\midrule
Temperature          & 0.6 \\
Top-$p$              & 0.95 \\
Top-$k$              & 20 \\
Max tokens           & 8192 \\
Judge model          & \texttt{gpt-5-mini} \\
\bottomrule
\end{tabular}
\caption{Evaluation hyperparameters.}
\label{tab:eval_settings}
\end{table}

\subsection{Licenses for existing assets}
\label{app:licenses}

All assets used in this work are publicly available and used in compliance with their respective licenses for non-commercial academic research. Qwen3-1.7B and Qwen3-8B~\citep{yang2025qwen3} are released under the Apache 2.0 License. WildChat~\citep{zhao2024wildchat} is released under ODC-BY 1.0. FollowBench~\citep{jiang2024followbench} is released under the Apache 2.0 License. AdvancedIF~\citep{he2025advancedif} is released under CC-BY-NC-4.0; we use it solely for evaluation in a non-commercial research context. InfoBench~\citep{qin2024infobench} is released under the MIT License. The gpt-5-mini judge is accessed through the OpenAI API in accordance with OpenAI's usage policies; we do not redistribute judge outputs.
The DeepSeek-V4-Pro and Gemini-3.5-Flash judges used in \aref{app:additional} are accessed through their providers' APIs in accordance with their usage policies.

\subsection{Compute resources}
\label{app:compute}

All experiments are run on NVIDIA H100 GPUs.
Each main result in \aref{tab:main_results} corresponds to a single training run on 4$\times$H100, taking 19.22\,h for LLM-as-a-Tutor end-to-end (rollout generation, tutor invocation, GRPO updates, and per-epoch tutor adaptation).
Baselines that share our GRPO setup (Base rubrics, WildChecklists, Policy-adaptive rubrics, EVA) consume comparable compute per run. Evaluation is run on the same hardware via vLLM~\cite{kwon2023efficient}, with judge calls served by the gpt-5-mini API; running all three benchmarks across five evaluation runs for one method costs approximately \$15 in API spend under OpenAI's flex tier.

\aref{tab:wallclock} compares the wall-clock time of our method with Base rubrics.
Tutor adaptation takes 36.19 min per epoch, or 1.81\,h of the 2.49\,h overhead over three epochs.

\begin{table}[h]
\centering
\small
\caption{Wall-clock training time of a full three-epoch run.}
\label{tab:wallclock}
\begin{tabular}{lccc}
\toprule
Method & Total training time & Additional overhead & Relative cost \\
\midrule
Base rubrics & 16.73\,h & -- & 100\% \\
LLM-as-a-Tutor & 19.22\,h & +2.49\,h & 115\% \\
\bottomrule
\end{tabular}
\end{table}

\section{Baseline details}\label{app:baseline_details}

\paragraph{SFT-distilled.} We fine-tune the Qwen3-1.7B policy on responses generated by the Qwen3-8B tutor for the same WildChat prompts used in our main experiments.
For each prompt, we sample a single response from Qwen3-8B and apply standard cross-entropy fine-tuning over the (prompt, tutor-response) pairs. Hyperparameters are summarized in \aref{tab:sft_distill_hparams}.

\begin{table}[h]
\centering
\small
\setlength{\tabcolsep}{6pt}
\captionsetup{skip=10pt}
\begin{tabular}{l l}
\toprule
Hyperparameter & Value \\
\midrule
Train batch size      & 16 \\
Learning rate         & 1e-5 \\
LR warmup ratio       & 0.05 \\
LR scheduler          & cosine \\
Epochs                & 3 \\
\bottomrule
\end{tabular}
\caption{SFT-distilled baseline hyperparameters. Sampling parameters are shared between the Qwen3-8B teacher and the Qwen3-1.7B student.}
\label{tab:sft_distill_hparams}
\end{table}

\paragraph{Evol-Instruct.}
Evol-Instruct~\citep{xu2023wizardlm} serves as our \emph{policy-agnostic} prompt adaptation baseline: prompts are rewritten offline using only the seed prompt itself, without consulting the policy.
Each WildChat seed prompt undergoes four sequential evolution rounds; at each round, we uniformly sample one of five mutation methods (four in-depth operators—\emph{Add Constraints}, \emph{Deepening}, \emph{Concretizing}, \emph{Increased Reasoning Steps}—and one in-breadth operator) and ask Qwen3-8B to rewrite the prompt under the corresponding template.\footnote{We use the mutation prompt templates from \texttt{distilabel}~v1.3.2 (\texttt{tasks/evol\_instruct/utils.py}).}
The output of the final round replaces the original prompt one-to-one.
Because our setup uses rubric-based RL, we then regenerate the base rubric on each rewritten prompt so that the reward signal stays aligned with the modified prompt.

\paragraph{EVA.}
EVA~\citep{ye2025evolving} selects prompts using a reward gap derived from a scalar reward model.
For each prompt $x$, responses $\{y_i\}_{i=1}^K$ sampled from the current policy are scored by a scalar reward model $r$, and the gap $\max_i r(x, y_i) - \min_i r(x, y_i)$ serves as the informativeness signal.
Each selected prompt is then completely rewritten using Evol-Instruct templates. The rewriting step itself does not condition on the sampled policy responses.

We re-implement EVA following the algorithmic specification in the original paper, with three deviations driven by our infrastructure and the rubric-based RL setup.
First, we train under GRPO rather than DPO~\citep{rafailov2023direct}.
Second, while the original paper uses \texttt{gemini-1.5-pro} to evolve prompts, we replace it with Qwen3-8B for consistency with the tutor used by our method and other baselines.
Third, the original paper derives the reward from a scalar reward model (\texttt{Skywork/Skywork-Reward-V2-Llama-3.1-8B}), whereas we use our rubric-based reward by regenerating the base rubric after each prompt evolution.
All other components, including the informativeness-based prompt selection and the prompt evolution templates, follow the original implementation.
These deviations are summarized in \aref{tab:eva_deviations}.

\begin{table}[h]
\centering
\small
\caption{Deviations between the original EVA specification and our re-implementation.}
\label{tab:eva_deviations}
\setlength{\tabcolsep}{6pt}
\begin{tabular}{l l l}
\toprule
Component & Original & Ours \\
\midrule
Algorithm       & DPO                       & GRPO \\
\makecell[l]{Prompt Evolver}  & \texttt{gemini-1.5-pro}   & Qwen3-8B \\
Reward model   & \makecell[l]{\texttt{Skywork-Reward-V2-Llama-3.1-8B}} & \makecell[l]{Qwen3-8B} \\
Reward signal       & scalar RM score                       & Rubric-based score \\
\bottomrule
\end{tabular}
\end{table}

\paragraph{Policy-adaptive rubric.}
We include policy-adaptive rubric generation as a baseline that adapts only the rubric, isolating the contribution of rubric-level adaptation from the prompt-level adaptation used in our main method.
As in LLM-as-a-Tutor, the tutor first judges whether the current rubric discriminates among the policy's rollouts; if not, it generates additional criteria targeting the specific weaknesses observed in those rollouts.
These adaptive criteria are appended to the base rubric at each training iteration; the prompt used for adaptive rubric generation is provided in \aref{app:adaptive_rubric_prompt}.
In contrast to \aref{eq:adapted}, which augments both the prompt and the rubric, here only the rubric is adapted while the prompt is left unchanged ($\tilde{x} = x$, $\tilde{R}(x) = R(x) \cup R_{adaptive}$).

\section{\texorpdfstring{Additional results and analyses}{Additional results and analyses}}\label{app:additional}
This section supplements \aref{sec:experiments}, \aref{sec:analysis}, and \aref{sec:discussion} with details regarding additional evaluations, baselines, and analyses of the tutor.

\subsection{\texorpdfstring{Main results with an alternative evaluation judge}{Main results with an alternative evaluation judge}}\label{app:add_main}
Because the tutor, the training-time judge, and the evaluation judge are all LLMs, we re-score the policies in \aref{tab:main_results} with DeepSeek-V4-Pro in place of GPT-5-mini as the evaluation judge.
All other evaluation settings follow \aref{app:evaluation}.
\aref{tab:deepseek_eval} shows that LLM-as-a-Tutor attains the highest average under DeepSeek-V4-Pro, as it does under GPT-5-mini, and outperforms all baselines on the same four of five metrics, with InfoBench DRFR as the exception.

\begin{table}[h]
\centering
\small
\caption{Benchmark scores of a subset of the policies in \protect\aref{tab:main_results}, judged by DeepSeek-V4-Pro instead of GPT-5-mini. Best results in each column are highlighted in bold. \textbf{LLM-as-a-Tutor achieves the best average under the alternative judge.}}
\label{tab:deepseek_eval}
\setlength{\tabcolsep}{4pt}
\begin{tabular}{lcccccc}
\toprule
& \multicolumn{2}{c}{FollowBench} & \multicolumn{2}{c}{AdvancedIF} & InfoBench & \\
Method & HSR & SSR & Overall & Micro & DRFR & Avg. \\
\midrule
Qwen3-1.7B & 41.60 & 62.27 & 15.22 & 67.90 & 73.28 & 52.05 \\
Base rubrics & 44.23 & 63.58 & 17.66 & 67.58 & \textbf{74.66} & 53.54 \\
Evol-Instruct & 43.68 & 63.41 & 16.32 & 67.29 & 73.85 & 52.91 \\
EVA & 46.24 & 64.67 & 17.61 & 68.29 & 73.99 & 54.16 \\
Policy-adaptive rubrics & 45.33 & 64.84 & 17.21 & 69.32 & 74.52 & 54.24 \\
LLM-as-a-Tutor & \textbf{46.79} & \textbf{65.83} & \textbf{18.31} & \textbf{70.08} & 74.32 & \textbf{55.07} \\
\bottomrule
\end{tabular}
\end{table}

\subsection{\texorpdfstring{Ablation on policy-adaptive prompt selection}{Ablation on policy-adaptive prompt selection}}\label{app:add_selection}
\aref{tab:rule_based} reports all five metrics for the two rule-based alternatives to our pairwise trigger in \aref{tab:adapt_ablation}.
Variance-based selection samples eight rollouts from the current policy, scores them on the base rubrics with the Qwen3-8B judge, and selects a prompt when the population standard deviation of these scores falls below $\tau=3$.
Similarity-based selection embeds the two policy rollouts with Qwen3-Embedding-8B and selects a prompt when their cosine similarity exceeds a threshold of 0.926.
We calibrate this threshold once on the base policy to match our constraint-added rate of 28.5\%.
A selected prompt receives one constraint from the Qwen3-8B tutor.
Like Always and Random, both rule-based baselines generate this constraint from the instruction alone, without the policy's rollouts.
Both alternatives improve over Base rubrics on average but fall short of our pairwise trigger, with the largest gaps on AdvancedIF.

\begin{table}[h]
\centering
\small
\caption{Comparison with rule-based prompt-selection baselines on all five metrics. \textbf{Pairwise judgments outperform both rule-based alternatives on average.}}
\label{tab:rule_based}
\setlength{\tabcolsep}{4pt}
\begin{tabular}{lcccccc}
\toprule
& \multicolumn{2}{c}{FollowBench} & \multicolumn{2}{c}{AdvancedIF} & InfoBench & \\
Method & HSR & SSR & Overall & Micro & DRFR & Avg. \\
\midrule
Qwen3-1.7B & 35.34 & 58.43 & 13.08 & 66.48 & 72.28 & 49.12 \\
Base rubrics & 38.60 & 60.17 & 13.98 & 66.32 & 73.48 & 50.51 \\
Similarity-based & 40.58 & 61.96 & 13.88 & 66.02 & 73.36 & 51.16 \\
Variance-based & 40.01 & 61.67 & 13.98 & 66.56 & \textbf{73.62} & 51.17 \\
LLM-as-a-Tutor & \textbf{40.91} & \textbf{62.28} & \textbf{15.07} & \textbf{67.97} & 73.59 & \textbf{51.96} \\
\bottomrule
\end{tabular}
\end{table}

\subsection{\texorpdfstring{Analysis of the tutor's behavior}{Analysis of the tutor's behavior}}\label{app:add_tutor}

\paragraph{Number of rollouts examined by the tutor.}
We vary the number of policy rollouts that the tutor examines per prompt over two, four, and eight, and compare the reward statistics of the prompts it selects and leaves unchanged as in \aref{fig:difficulty_selection}.
\aref{tab:rollout_count} shows a similar qualitative separation for every setting, while the adaptation time per epoch grows in proportion to the number of rollouts.

\begin{table}[h]
\centering
\small
\caption{Ablation on the number of rollouts examined by the tutor. Selected prompts are those the tutor identifies as non-challenging, and reward std is the mean per-prompt population standard deviation of base-rubric reward over eight rollouts as in \protect\aref{fig:difficulty_selection}. For four and eight rollouts, the selection statistics are computed on a 1,000-prompt subsample. \textbf{Two rollouts give a similar separation at half or a quarter of the adaptation time.}}
\label{tab:rollout_count}
\setlength{\tabcolsep}{4pt}
\begin{tabular}{lccccc}
\toprule
& \multicolumn{2}{c}{Selected} & \multicolumn{2}{c}{Unselected} & \\
\# Rollouts & Mean & Reward std & Mean & Reward std & Adaptation time (1 epoch) \\
\midrule
2 & 90.76 & 6.15 & 78.24 & 10.04 & 36.19 min ($\times$1.00) \\
4 & 90.25 & 6.51 & 77.68 & 11.02 & 72.67 min ($\times$2.01) \\
8 & 90.32 & 6.55 & 76.25 & 11.50 & 143.05 min ($\times$3.95) \\
\bottomrule
\end{tabular}
\end{table}

\paragraph{Constraint quality.}
We fix the policy to Qwen3-1.7B and vary the tutor (Qwen3-8B, Qwen3-32B, and DeepSeek-V4-Pro) and the judge (GPT-5-mini and Gemini-3.5-Flash).
Each judge evaluates 441, 215, and 242 constraint-addition events from Qwen3-8B, Qwen3-32B, and DeepSeek-V4-Pro, respectively, sampled from the first three epochs.
The judge sees the seed instruction, the previously appended constraints, and the new constraint, and gives a pass or violate verdict on each dimension with the prompt in \aref{app:quality_judge_prompt}.
\textbf{Compatibility} means that a response can jointly satisfy the new constraint, all prior constraints, and the original requirements of the seed instruction.
\textbf{Groundedness} means that the constraint is factually correct and refers only to entities present in the seed instruction and prior constraints.
\textbf{Self-verifiability} means that compliance can be verified from the instruction and response alone without external knowledge.
\textbf{Intent preservation} means that the constraint maintains the original intent.
This groundedness criterion is stricter than the generation prompt, which also allows references to the rollouts.
Qwen3-8B and Qwen3-32B tutors are comparable on every dimension in \aref{tab:constraint_quality}.

\begin{table}[h]
\centering
\small
\caption{Quality evaluation of generated constraints across tutor models, rated by two judges. \textbf{Constraints from every tutor are compatible and preserve the original intent.}}
\label{tab:constraint_quality}
\setlength{\tabcolsep}{4pt}
\begin{tabular}{llcccc}
\toprule
Tutor & Judge & Compatibility & Groundedness & \makecell{Self-\\verifiability} & \makecell{Intent\\preservation} \\
\midrule
Qwen3-8B & GPT-5-mini & 99.7\% & 97.1\% & 90.7\% & 99.5\% \\
Qwen3-8B & Gemini-3.5-Flash & 98.8\% & 93.9\% & 93.6\% & 96.6\% \\
Qwen3-32B & GPT-5-mini & 98.3\% & 96.3\% & 91.6\% & 98.6\% \\
Qwen3-32B & Gemini-3.5-Flash & 99.1\% & 94.4\% & 93.0\% & 96.3\% \\
DeepSeek-V4-Pro & GPT-5-mini & 100.0\% & 98.8\% & 97.1\% & 100.0\% \\
DeepSeek-V4-Pro & Gemini-3.5-Flash & 100.0\% & 99.2\% & 100.0\% & 98.8\% \\
\bottomrule
\end{tabular}
\end{table}

\subsection{\texorpdfstring{Generalizing beyond constraint-additive tasks}{Generalizing beyond constraint-additive tasks}}\label{app:add_generalize}
Our FollowBench subset spans 620 instances across 13 task families, listed in \aref{tab:followbench_tasks}.
We also evaluate on Creative Writing v3 with the official creative-writing-bench defaults, 32 prompts with three iterations each (96 items), and report the instruction-adherence sub-score judged by Claude Sonnet 4.6.
LLM-as-a-Tutor scores 5.38, compared with 5.21 for Base rubrics and 4.98 for the untrained Qwen3-1.7B.

\begin{table}[h]
\centering
\small
\caption{Task distribution in FollowBench, for the subset we evaluate.}
\label{tab:followbench_tasks}
\begin{tabular}{lrr}
\toprule
Task & \#Data & Ratio \\
\midrule
Open-ended Question Answering & 295 & 47.6\% \\
Complex Situation Reasoning & 55 & 8.9\% \\
Suggestion Generation & 40 & 6.5\% \\
Text-to-Table Generation & 30 & 4.8\% \\
Data-to-Text Generation & 25 & 4.0\% \\
Document-Level Event Argument Extraction & 25 & 4.0\% \\
Document-Level Named Entity Recognition & 25 & 4.0\% \\
Text Generation with Language Constraints & 25 & 4.0\% \\
Text Editing & 25 & 4.0\% \\
Summarization & 25 & 4.0\% \\
Machine Translation & 25 & 4.0\% \\
Role-playing & 15 & 2.4\% \\
Story Generation & 10 & 1.6\% \\
\midrule
Total & 620 & 100\% \\
\bottomrule
\end{tabular}
\end{table}

\subsection{\texorpdfstring{Prompt growth and constraint removal}{Prompt growth and constraint removal}}\label{app:add_growth}
\aref{tab:append_dist} counts the constraints appended to each of the 3,994 training prompts over the three epochs.
The tutor makes one decision per prompt in each epoch, so three is the most a prompt can receive.
The 3,322 appends over three epochs of 3,994 prompts give the three-epoch average constraint-added rate of 27.7\%, the 28\% in \aref{subsec:ablation_select}, while the tutor's first pass adds constraints to 25.8\% of prompts (\aref{fig:constraint_density}).

\begin{table}[h]
\centering
\small
\caption{Number of constraints appended per prompt over 3 epochs.}
\label{tab:append_dist}
\begin{tabular}{lrr}
\toprule
Number of appends & Prompts & Percentage \\
\midrule
0 & 1,851 & 46.34\% \\
1 & 1,205 & 30.17\% \\
2 & 697 & 17.45\% \\
3 & 241 & 6.03\% \\
\bottomrule
\end{tabular}
\end{table}

\clearpage
\section{Prompt templates}\label{app:prompt_templates}
This section lists the prompt templates used in our pipeline.
Slots wrapped in \texttt{\{\{...\}\}} are filled at runtime with the corresponding field.

\subsection{Judge prompt}\label{app:judge_prompt}
Judge prompt is adapted from RLCF~\citep{viswanathan2025checklists}, with minor modifications. The full template is shown below.
\begin{promptbox}[Judge prompt]
\begin{verbatim}
You are evaluating a model's response to an instruction against a specific criterion. Your task is to score how well the response satisfies that criterion, given the context of the instruction.

Score the response on a scale of 0 to 100 using the following anchors. For example, the instruction might be "What is a good vegan substitute to meat for someone allergic to soy and gluten? Provide a single-sentence response consisting of an answer followed by a factually detailed and humorous one-sentence explanation", and the criterion might be "Is the explanation factually detailed?". Your selection should be based on the response and the criterion, using the following rating scale:

- 100: The response represents an optimal solution that expertly balances all relevant aspects of the instruction. For the example above (about the vegan substitute), and the criterion above (about factual detail), an example 100-point response is "Mushrooms, because they can be easily caramelized and browned, they are rich in the glutamates which lead to incredible umami flavors, they naturally are completely free of soy and gluten, and they don't look cute as babies". This response is richly detailed and factual, and though it fails to be humorous, it is still a 100-point response on the factual detail criterion.

- 75: The response very effectively addresses the criterion but has room for minor improvements. The response should be unconditionally acceptable (at a professional level) but may not be absolutely perfect. There are no mistakes that critically undermine the criterion. An example 75-point response to the example criterion above is "Mushrooms - they are rich in the glutamates that lead to incredible umami flavors and they don't look cute in the slightest while alive.". This response has one interesting fact but could be more detailed.

- 50: The response adequately fulfills the criterion but contains notable flaws or missed opportunities for improvement. The response should still be functionally acceptable. The response contains at most one minor inadequacy or inaccuracy related to the criterion but there are no mistakes that critically undermine the criterion. An example 50-point response to the example criterion above is "Mushrooms, because they can be easily caramelized and browned, they're universally beloved by sophisticated palates, and they don't look cute in the slightest while alive." The statement that they're universally beloved by people with sophisticated palates, while potentially true, is vague and not objective.

- 25: The response fulfills the key condition specified by the criterion and demonstrates awareness of the key condition but fails to execute them effectively. The text may contain non-critical inaccuracies or irrelevant information. However, if there is even one element that critically undermines the core purpose specified in the criterion (even if that element seems minor in isolation), the score should be 0 (not 25). An example 25-point response to the example criterion above is "Mushrooms, because they can be easily caramelized and browned, they are absolutely brimming with protein, and they don't look cute in the slightest while alive." The claim that mushrooms are "absolutely brimming with protein" is factually inaccurate.

- 0: The response fails to meet the criterion or provides no information that could be utilized to answer the criterion. If the response contains a critical error relevant to the criterion, return a 0. For the criterion about the vegan substitute, an example 0-point response is "Mushrooms, because they make you question why you ever thought a dead animal could compare to this vegan delight." While funny and engaging, this response contains zero factual detail about mushrooms, critically violating the criterion.

Your score can be any integer between 0 and 100 (not just the ones listed above). If you are totally confused, return -1.

Evaluate the response ONLY against the stated criterion. Do not penalize or reward aspects of the response unrelated to it.

<instruction>
{{instruction}}
</instruction>

<response>
{{response}}
</response>

<criterion>
{{requirement}}
</criterion>

After examining the instruction, the response, and the criterion:

- Briefly justify your score in up to 100 words.
- Conclude with the score using the following format:
  <score>{integer between 0 and 100 or -1}</score>
\end{verbatim}
\end{promptbox}

\subsection{Tutor prompts}\label{app:tutor_prompt}

\subsubsection{Base rubric generation}\label{app:base_rubric_prompt}
Given a seed prompt, the tutor generates a list of weighted criteria that constitute the base rubric.
\begin{promptbox}[Base rubric generation prompt]
\begin{verbatim}
You are responsible for developing criteria for judging arbitrary responses to instructions. You will be given an instruction, and your task is to write a set of rubric criteria for evaluating responses to the instruction.

---

# Task Guidelines

## Step 1 - Analyze the instruction

Analyze the instruction to understand the key quality dimensions needed to evaluate potential responses to the instruction. Identify the task scope and the explicit constraints specified in the instruction.

## Step 2 - Write rubric criteria

Write the rubric criteria in the form of complete questions. Example: 'Does the response include at least three examples?'

Each criterion question must follow these four principles:
- **Self-contained**: The criterion can be evaluated independently, without reading other criteria or needing extra context beyond the instruction and response.
- **Atomic**: Each criterion judges exactly one thing. Do not bundle multiple requirements into a single criterion.
- **Non-redundant**: No two criteria overlap or measure the same underlying quality.
- **Diverse**: Collectively, the criteria should cover a range of dimensions (e.g., correctness, depth, clarity, structure, style) rather than clustering on one aspect.

For each criterion question, assign an **importance score** between 0 and 100.
- 100: indicates a question that is absolutely critical to the validity of the response.
- 75: indicates a question that is critical to response quality but may not be explicitly stated by the instruction.
- 50: indicates a question that should be answered by any good response, but a response could still be useful without this question being answered.
- 25: indicates a question that is a preference but not a requirement.
- Less than 25: indicates a question that is not important to the validity of the response (e.g. a soft nice-to-have).

---

# Input

## Instruction

<instruction>
{{instruction}}
</instruction>

---

# Output Format

<analysis>
Summary of your analysis of the instruction.
</analysis>
<rubric>
    <criterion>Criterion question 1</criterion>
    <importance>integer score between 0 and 100</importance>
    ...
    <criterion>Criterion question N</criterion>
    <importance>integer score between 0 and 100</importance>
</rubric>
\end{verbatim}
\end{promptbox}

\subsubsection{Constraint generation prompt}\label{app:constraint_prompt}
Given a prompt and a pair of policy rollouts, the tutor decides whether the prompt is saturated for the current policy; if so, it appends a single targeted constraint.
The decision, the constraint, and its rubric criteria come from this single call, and the \texttt{\{\{instruction\}\}} slot holds the current accumulated prompt $x^{(t)}$, including previously appended constraints.
\begin{promptbox}[Constraint generation prompt]
\begin{verbatim}
You are evaluating an instruction and two model responses. Your task is to determine whether the model has completely mastered this instruction -- leaving no room for quality improvement -- and therefore needs an additional constraint to create a meaningful learning signal.

The default answer is NO. Only say YES when the evidence clearly shows the model has hit its quality ceiling on this instruction.

---

# Step 1 - Evaluate each response independently

For each response, assess:
**Accuracy**: Is the information correct? Are there any factual errors, hallucinations, unsupported claims, or uncertain guesses?
**Completeness**: Does it address everything the instruction asks for? Is anything missing or treated superficially?
**Depth**: Does it go beyond surface-level treatment? Would a knowledgeable reader find it thorough, or would they want more?
**Execution quality**: Is it well-organized, clear, and appropriate in tone for the task?

Summarize the quality of each response. Be honest and critical -- look for genuine weaknesses, not just things you could add.

# Step 2 - Compare the two responses

Now compare them:

**Quality gap**: Is one response noticeably better than the other? If so, the model produces variable quality on this instruction -- there is already a gradient to learn from.
-> `<decision>no</decision>`
**Approach divergence**: Do the responses take meaningfully different approaches, structures, or perspectives? If so, the model is still exploring how best to handle this instruction -- variance already exists.
-> `<decision>no</decision>`
**Content convergence**: Do both responses arrive at essentially the same answer with the same structure and coverage? If so, this is a sign the model has settled on one approach and produces it consistently.

# Step 3 - Make the decision

Say YES only when ALL of the following are true:
- Both responses are **high quality** -- accurate, complete, well-executed. Neither has meaningful weaknesses.
- Both responses **converge** -- similar approach, structure, and content. They are nearly interchangeable.
- There is **no gradient** -- you cannot meaningfully say one is better than the other, because both are equally strong.

If ANY of the following is true, say NO:
- Either response has factual errors, hallucinations, or incorrect information.
- Either response is superficial, generic, or lacks depth on a topic that warrants it.
- Either response misses something the instruction asked for.
- One response is clearly better than the other.
- The responses take substantially different approaches or cover different ground.
- Either response shows uncertainty (hedging, listing multiple possibilities, "it could be...").

When in doubt, say NO.

# Step 4 - Write the constraint (only if decision is yes)

## 4a. Identify what to target

Both responses do everything well -- what dimension is entirely absent? Target only something neither response attempted.

## 4b. Choose a dimension

Select ONE:
- **Content**: Modify scope, add subtasks, mandate components.
- **Situation**: Impose a scenario, condition, edge case, or role.
- **Style**: Require a specific tone, voice, or persona.
- **Format**: Require a specific structure, length limit, or presentation style.

## 4c. Write the constraint

Write a single, atomic constraint that is:

- **Challenging**: Targets something neither response attempted.
- **Natural**: Reads as a seamless continuation of the instruction. Uses a connector (e.g., "Additionally,"). No meta-language.
- **Additive**: A genuinely new demand, not a restatement of anything already in the instruction.
- **Specific**: Concrete and evaluable.
- **Factually grounded**: Only references things present in the instruction or responses.
- **Atomic**: One requirement only.

The constraint must NOT be a generic filler ("include an example", "add a case study") or a quality baseline ("ensure accuracy").

# Step 5 - Write rubric criteria for the constraint
# (only if decision is yes)

Write rubric criteria that cover ONLY the new constraint from Step 4 -- not the original instruction. Write **at most 2 criteria**. If the constraint can be fully captured in 1 criterion, use only 1.

Each criterion must be a complete question (e.g., "Does the response include at least three examples?") and follow these principles:
- **Self-contained**: Evaluable independently, without needing other criteria or extra context.
- **Atomic**: Judges exactly one thing.
- **Positively framed**: A "yes" answer means the response satisfies the criterion.

For each criterion, assign an **importance score** (0-100):
- 100 = absolutely critical to validity.
- 75 = critical to quality but not explicitly stated.
- 50 = expected of any good response.
- 25 = preference, not a requirement.

---

# Input

## Instruction

<instruction>
{{instruction}}
</instruction>

## Model Responses

<response_1>
{{response_1}}
</response_1>

<response_2>
{{response_2}}
</response_2>

---

# Output Format

<analysis>
Step 1: Quality assessment of each response -- strengths and genuine weaknesses.
Step 2: Comparison -- quality gap? approach divergence? content convergence?
Step 3: Decision -- has the model fully mastered this instruction with no room to improve?
Step 4 (if yes): What is absent? Chosen dimension and rationale.
Step 5 (if yes): Rubric criteria for the constraint.
</analysis>
<decision>yes or no</decision>
<constraint>
The single constraint to append (required only when decision is yes).
</constraint>
<rubric>
    <criterion>Criterion question 1</criterion>
    <importance>integer score between 0 and 100</importance>
    <criterion>Criterion question 2 (optional)</criterion>
    <importance>integer score between 0 and 100</importance>
</rubric>
\end{verbatim}
\end{promptbox}

\subsubsection{Policy-adaptive rubric generation}\label{app:adaptive_rubric_prompt}
\begin{promptbox}[Policy-adaptive rubric generation prompt]
\begin{verbatim}
You are evaluating an instruction, two model responses, and an existing rubric. Your task is to determine whether the existing rubric fails to discern the two responses -- catching none of the observable failures -- and therefore needs an additional criterion. The default is NO; say YES only when the evidence clearly shows the rubric cannot discern the responses.

---

# Step 1 - Spot concrete, observed failures

For each response, list specific, instruction-relevant **failures** -- missing required elements, internal contradictions, malformed structure, misinterpretations, obvious quality defects -- that you can verify directly from the text. Tag which response(s) exhibit each.

**Self-verifiability (hard constraint).** Only count failures confirmable from the response text and the instruction alone -- not anything requiring external knowledge, factual lookup, or domain expertise (e.g., "the cited paper does not exist", "the historical date is wrong"). If unsure, drop.

If neither response has a self-verifiable failure, the decision will be NO.

# Step 2 - Compare against the existing rubric

**Existing gradient**: Do one or more existing criteria already produce a different answer on response 1 vs response 2? If so, the rubric is already discriminating -- there is already signal to learn from.
-> `<decision>no</decision>`

**Any covered failure**: For each failure from Step 1, ask whether an existing criterion -- even approximately -- would catch it. If **at least one** is already covered, the rubric is doing meaningful work. Leave it alone.
-> `<decision>no</decision>`

**No coverage at all**: Does the rubric fail to catch a single observed failure, and produce no differentiating signal between the two responses? Only this case warrants a new criterion.

# Step 3 - Make the decision

Say YES only when ALL are true:
- At least one concrete, self-verifiable failure identified.
- **None** of the failures is caught -- even approximately -- by any existing criterion.
- **No existing criterion** produces a different answer between the two responses.
- The uncaught failure(s) represent a class of problem any plausible response could exhibit -- not a quirk of these samples.

Say NO if ANY is true (when in doubt, say NO):
- No self-verifiable failures found.
- At least one failure is already caught (or nearly caught) by an existing criterion.
- Existing rubric already produces a different answer on at least one criterion.
- Failure requires external knowledge, or is specific to these samples rather than a class of problem.
- Uncertain whether any criterion catches any failure.

# Step 4 - Write new criteria (only if decision is yes)

## 4a. Select targets

From the uncovered failures, select the most load-bearing class(es). Target **at most 2 criteria**; use 1 if sufficient.

## 4b. Generalize each failure into a criterion

For each selected failure, write **one criterion question** that detects this *class* of problem in **any** response to this instruction.

Each criterion must satisfy ALL of these:

- **Grounded.** Cites a specific failure from Step 1. If you cannot cite one, drop it.
- **Substantive, not surface.** Targets a meaningful quality dimension (content, reasoning, coverage, depth, accuracy) -- not formatting, typography, specific word choice, or length counts. Ask what underlying quality the failure manifests and write around that. Do **not** write criteria like "Does the response use bullet points?", "Does the response include 'therefore'?", or "Is the response at least 500 words?" unless the instruction demands such form.
- **Generalized.** Does not reference entities or phrases appearing **only** in the sample responses. *Exception:* items the instruction itself enumerates (named characters, required sections, listed keywords) may be referenced by name.
- **Self-contained.** Evaluable from the instruction and a single response alone.
- **Atomic.** Judges exactly one thing.
- **Reachable.** A competent, good-faith response could reasonably satisfy it. No extreme demands ("lists 50 examples", "cites 10 peer-reviewed sources") unless the instruction requires them.
- **Natural to the task type.** Any structural or formal demand must appear in the instruction or be a natural expectation of the task type (a story has a narrative arc; a financial report shows numbers). Do **not** impose structure the instruction does not require and the task does not naturally imply.
- **Non-trivial.** Could plausibly separate a good response from a mediocre one. "Any reasonable attempt passes" is too weak -- avoid vague checks like "addresses the topic" or "is in English" that any competent response would pass.
- **Self-verifiable.** A judge reading only the instruction and one response can score it without external knowledge.
- **Complete question.** E.g., "Does the response include a clearly labeled summary section?"
- **Positive polarity.** Higher score clearly means a better response. Avoid "A or B" listings (bad: *"...list only beginner exercises, or include advanced variations?"*; good: *"...restrict its list to beginner exercises?"*). When the instruction forbids X, phrase around successfully avoiding X (bad: *"Does the response use passive voice?"*; good: *"Does the response avoid passive voice throughout?"*).

## 4c. Assign importance (0-100)

**100** critical to validity * **75** critical to quality, not explicit in instruction * **50** expected of any good response * **25** preference * **<25** soft nice-to-have.

---

# Input

## Instruction

<instruction>
{{instruction}}
</instruction>

## Existing Rubric

<existing_rubric>
{{existing_rubric}}
</existing_rubric>

## Model Responses

<response_1>
{{response_1}}
</response_1>

<response_2>
{{response_2}}
</response_2>

---

# Output Format

<analysis>
Step 1: Self-verifiable failures, tagged with which response(s) exhibit them.
Step 2: For each failure, whether an existing criterion covers it; whether any criterion already discriminates.
Step 3: Decision -- is it difficult to discern the responses using the existing rubric?
Step 4 (if yes): For each new criterion, name the underlying failure and confirm it is substantive, natural to the task type, non-trivial, reachable, self-verifiable, and generalized.
</analysis>
<decision>yes or no</decision>
<rubric>
    <criterion>Criterion question 1</criterion>
    <importance>integer score between 0 and 100</importance>
    <criterion>Criterion question 2 (optional)</criterion>
    <importance>integer score between 0 and 100</importance>
</rubric>
\end{verbatim}
\end{promptbox}

\subsubsection{Rewrite variant of LLM-as-a-Tutor}\label{app:rewrite_prompt}
\begin{promptbox}[Rewriting version of constraint generation prompt]
\begin{verbatim}

You are evaluating an instruction and two model responses. Your task is to determine whether the model has completely mastered this instruction -- leaving no room for quality improvement -- and therefore needs more challenging instruction to create a meaningful learning signal. If needed, you will REWRITE the original instruction so that the new instruction is asking for more difficult tasks.

The default answer is NO. Only say YES when the evidence clearly shows the model has hit its quality ceiling on this instruction.

---

# Step 1 -- Evaluate each response independently

For each response, assess:

**Accuracy**: Is the information correct? Are there any factual errors, hallucinations, unsupported claims, or uncertain guesses?

**Completeness**: Does it address everything the instruction asks for? Is anything missing or treated superficially?

**Depth**: Does it go beyond surface-level treatment? Would a knowledgeable reader find it thorough, or would they want more?

**Execution quality**: Is it well-organized, clear, and appropriate in tone for the task?

Summarize the quality of each response. Be honest and critical -- look for genuine weaknesses, not just things you could add.

# Step 2 -- Compare the two responses

Now compare them:

**Quality gap**: Is one response noticeably better than the other? If so, the model produces variable quality on this instruction -- there is already a gradient to learn from. -> `<decision>no</decision>`

**Approach divergence**: Do the responses take meaningfully different approaches, structures, or perspectives? If so, the model is still exploring how best to handle this instruction -- variance already exists. -> `<decision>no</decision>`

**Content convergence**: Do both responses arrive at essentially the same answer with the same structure and coverage? If so, this is a sign the model has settled on one approach and produces it consistently.

# Step 3 -- Make the decision

Say YES only when ALL of the following are true:
- Both responses are **high quality** -- accurate, complete, well-executed. Neither has meaningful weaknesses.
- Both responses **converge** -- similar approach, structure, and content. They are nearly interchangeable.
- There is **no gradient** -- you cannot meaningfully say one is better than the other, because both are equally strong.

If ANY of the following is true, say NO:
- Either response has factual errors, hallucinations, or incorrect information.
- Either response is superficial, generic, or lacks depth on a topic that warrants it.
- Either response misses something the instruction asked for.
- One response is clearly better than the other.
- The responses take substantially different approaches or cover different ground.
- Either response shows uncertainty (hedging, listing multiple possibilities, "it could be...").

When in doubt, say NO.

# Step 4 -- Identify the dimension to push (only if decision is yes)

## 4a. Pick the dimension

Both responses do everything well -- along which dimension is the task being satisfied too easily? Target the dimension where asking for more would force the model past its current ceiling.

## 4b. Choose a dimension

Select ONE:
- **Content**: Tighten scope, deepen coverage, or demand more rigorous subject matter.
- **Situation**: Constrain the scenario, condition, edge case, or role.
- **Style**: Demand a specific tone, voice, or persona.
- **Format**: Demand a specific structure, length limit, or presentation style.

The harder version must be:
- **Challenging**: Pushes the dimension past what either response attempted.
- **Specific**: Concrete and evaluable.
- **Factually grounded**: Only references things present in the instruction or responses.

# Step 5 -- Rewrite the instruction (only if decision is yes)

Rewrite the instruction over from scratch, asking for a harder version of the same task pushed along the dimension you chose in Step 4.

If decision is no, leave the `<rewrite>` block empty.

---

# Input

## Instruction

<instruction>
{instruction}
</instruction>

## Model Responses

<response_1>
{response_1}
</response_1>

<response_2>
{response_2}
</response_2>

---

# Output Format

<analysis>
Step 1: Quality assessment of each response -- strengths and genuine weaknesses.
Step 2: Comparison -- quality gap? approach divergence? content convergence?
Step 3: Decision -- has the model fully mastered this instruction with no room to improve?
Step 4 (if yes): Which dimension is being satisfied too easily? Chosen dimension and how the rewrite will push it.
Step 5 (if yes): Brief note on the rewrite approach.
</analysis>
<decision>yes or no</decision>
<rewrite>
The full rewritten instruction (required only when decision is yes). It is the same task asked in a harder form along the chosen dimension, written from scratch as one coherent prompt.
</rewrite>

\end{verbatim}
\end{promptbox}
\subsection{\texorpdfstring{Constraint-quality judge prompt}{Constraint-quality judge prompt}}\label{app:quality_judge_prompt}
The constraint-quality study in \aref{app:add_tutor} uses this prompt for both judges.
\begin{promptbox}[Constraint-quality judge prompt]
\begin{verbatim}
You are auditing a constraint that a tutor model appended to a training
instruction. The tutor examines pairs of responses that a policy model
produces for an instruction. When the responses no longer differ in quality,
the instruction has lost its discriminative power, and the tutor appends one
constraint to restore it. A valid constraint raises difficulty without
breaking the instruction it extends.

Your task is to judge whether this addition is valid along four independent
dimensions: Compatibility, Groundedness, Self-verifiability, and Intent
preservation. Judge each dimension independently. An addition can violate
several dimensions at once, and a violation on one dimension must not
influence your judgment on another.

---

# Step 1 - Compatibility

**Question**: Can a single response satisfy the new constraint and all
previously appended constraints and the seed instruction at the same time?

Compare the new constraint against the previously appended constraints and
the seed instruction. Mark **violate** when no response can satisfy them
together, whether because the demands are logically contradictory or because
they cannot all be met within the limits the instruction sets.

A constraint that makes the task harder but remains jointly satisfiable with
the previous requirements is a **pass**. Increased difficulty is the intended
effect of a constraint, not a defect.

# Step 2 - Groundedness

**Question**: Is the new constraint factually correct and a valid demand?

Mark **violate** in either of these cases:
- The constraint states or presupposes something that is false, or that
  contradicts what the seed instruction or the previous constraints establish.
- The constraint refers to material that appears nowhere in the seed
  instruction or the previous constraints.

Mark **violate** only for errors you can identify with confidence. Do not mark
violate merely because you cannot confirm a claim.

# Step 3 - Self-verifiability

**Question**: Is the constraint self-contained? Can a judge who reads only the
accumulated instruction and a single response determine whether the response
satisfies it, without external or domain knowledge?

Mark **pass** when the constraint requires nothing outside the response
itself, when the accumulated instruction already supplies what it refers to,
or when ordinary general knowledge settles it. Mark **violate** only when
compliance could not be settled without looking up a referent, or without
knowledge of a particular work, place, record, or specialty lying well beyond
general knowledge.

Judge what the response would have to reach for, not how firmly the constraint
is worded: a detail the response is free to invent stays self-contained
however emphatically it is demanded.

# Step 4 - Intent preservation

**Question**: Assuming a response satisfies the new constraint, is that
response still the kind of output the seed instruction asked for?

Compare the new constraint against the seed instruction. Mark **violate**
when satisfying the constraint changes the nature of the task. A constraint
that narrows, deepens, or specializes the original task while keeping its
purpose and output type intact is a **pass**.

# Step 5 - Verdicts

Mark a verdict for each dimension with a one-sentence justification
citing specific text from the materials.

---

# Input

## Seed Instruction
<instruction>
{{instruction}}
</instruction>

## Previously Appended Constraints
<previous_constraints>
{{previous_constraints}}
</previous_constraints>

## New Constraint Under Audit
<constraint>
{{constraint}}
</constraint>

---

# Output Format

<analysis>
Step 1: Compatibility. Can the new constraint and the previous requirements be satisfied at once?
Step 2: Groundedness. Does the constraint contain a factual error or an invalid reference?
Step 3: Self-verifiability. Can a judge check compliance from the accumulated instruction and a single response alone?
Step 4: Intent preservation. Does a satisfying response remain the task the seed asked for?
</analysis>
<compatibility>pass or violate</compatibility>
<groundedness>pass or violate</groundedness>
<self_verifiability>pass or violate</self_verifiability>
<intent_preservation>pass or violate</intent_preservation>
\end{verbatim}
\end{promptbox}

\newpage

\section{Additional qualitative examples}
\label{app:additional_qualitative}

\begin{figure}[!ht]

\small
\setlength{\tabcolsep}{8pt}
\renewcommand{\arraystretch}{1.25}
\begin{tabular}{p{0.45\textwidth} | p{0.45\textwidth}}
\toprule
\textbf{Before} ($x$) & \textbf{After} ($\tilde{x} = x \oplus c$) \\
\midrule
\textbf{Prompt:} Define ``humanistic techniques'' and explain how \texttt{TPR}, \texttt{Silent Way}, \texttt{CLL}, \texttt{Suggestopedia}, and \texttt{Natural Approach} illustrate the principles.
&
\textbf{Prompt:} \textit{\{Original Prompt\}} + \emph{``Additionally, \textcolor{red}{cite the originators of each technique} and briefly discuss their contributions to the development of humanistic techniques.''} \\
\midrule
$\mu_{\text{base}} = 100.0$, \quad $\sigma_{\text{base}} = 0.0$ & $\mu_{\text{base}} = 80.7$, \quad $\sigma_{\text{base}} = 14.7$ \\
\midrule
\multicolumn{2}{l}{\textbf{Sample rollouts on the adapted prompt} ($\tilde{x}$):} \\
\midrule
\multicolumn{2}{l}{\parbox{0.94\textwidth}{\vspace{2pt}
\textcolor{green!60!black}{\ding{52}} \textbf{Rollout A} ($s_{\text{base}}=96.1$):

1. \textbf{Total Physical Response (TPR)} \textit{[...]} \textbf{Originator}: \textcolor{red}{Peter Smith} (1960s).

2. \textbf{The Silent Way} \textit{[...]} \textbf{Originator}: Caleb Gattegno (1950s).

3. \textbf{Community Language Learning (CLL)} \textit{[...]} \textbf{Originator}: \textcolor{red}{John C. W. Kachru} (1960s).

4. \textbf{Suggestopedia} \textit{[...]} \textbf{Originator}: \textcolor{red}{Joseph K. Kachru} (1950s).

5. \textbf{The Natural Approach} \textit{[...]} \textbf{Originator}: Stephen Krashen (1980s).
\vspace{2pt}}} \\
\midrule
\multicolumn{2}{l}{\parbox{0.94\textwidth}{\vspace{2pt}
\textcolor{red}{\ding{56}} \textbf{Rollout B} ($s_{\text{base}}=81.1$):

1. \textbf{Total Physical Response (TPR)} \textit{[...]} \textbf{Originator}: \textcolor{red}{Robert L. Lado} (1963)

2. \textbf{The Silent Way} \textit{[...]} \textbf{Originator}: \textcolor{red}{James A. Smith} (1960s)

3. \textbf{Community Language Learning} \textit{[...]} \textbf{Originator}: \textcolor{red}{John L. S.} (1960s) -- \textbf{Contribution}: \textcolor{red}{Suggestopedia, developed by Ivan Pavlov}, was a precursor, but Community Language Learning emphasizes the role of the community in language acquisition \textit{[...]}

4. \textbf{Suggestopedia} \textit{[...]} \textbf{Originator}: \textcolor{red}{Ivan Pavlov} (1930s)

5. \textbf{The Natural Approach} \textit{[...]} \textbf{Originator}: Stephen Krashen (1985)
\vspace{2pt}}} \\
\bottomrule
\end{tabular}
\caption{Qualitative example: defining humanistic techniques and walking through the five named methods is a task on which the policy's base rollouts saturate at the ceiling ($\sigma_{\text{base}} = 0$). Once the constraint asks each technique's originator to be named, the rollouts hallucinate to differing degrees: Rollout A misattributes some methods but keeps correct anchors such as Gattegno for the Silent Way, while Rollout B fabricates throughout and compounds the error by attributing Suggestopedia to Ivan Pavlov inside a contribution sentence.}
\label{fig:qualitative_humanistic_teaching}
\end{figure}

\begin{figure}[!t]

\small
\setlength{\tabcolsep}{8pt}
\renewcommand{\arraystretch}{1.25}
\begin{tabular}{p{0.45\textwidth} | p{0.45\textwidth}}
\toprule
\textbf{Before} ($x$) & \textbf{After} ($\tilde{x} = x \oplus c$) \\
\midrule
\textbf{Prompt:} How should the Acharjee et al.\ (2018) citation be formatted as an \texttt{OSCOLA} footnote vs.\ bibliography entry, with example.
&
\textbf{Prompt:} \textit{\{Original Prompt\}} + \emph{``Additionally, ensure the response \textcolor{red}{specifies whether the title and journal name in the bibliography entry should be italicized} and includes the exact punctuation rules for the footnote format.''} \\
\midrule
$\mu_{\text{base}} = 95.3$, \quad $\sigma_{\text{base}} = 7.6$ & $\mu_{\text{base}} = 84.7$, \quad $\sigma_{\text{base}} = 14.7$ \\
\midrule
\multicolumn{2}{l}{\textbf{Sample rollouts on the adapted prompt} ($\tilde{x}$):} \\
\midrule
\multicolumn{2}{l}{\parbox{0.94\textwidth}{\vspace{2pt}
\textcolor{green!60!black}{\ding{52}} \textbf{Rollout A} ($s_{\text{base}}=99.2$):

\textbf{Punctuation rules:}
\begin{itemize}\setlength{\itemsep}{0pt}
\item \textbf{Author(s)}: Full names, separated by commas.
\item \textbf{Year}: In parentheses (e.g., 2018).
\item \textbf{Title}: Italicized (e.g., \emph{Electronic waste management...}).
\item \textbf{Journal name}: Italicized (e.g., \emph{Journal of Environmental Treatment Techniques}).
\end{itemize}
\textit{[...]} \textbf{Title and journal name}: Always \textbf{italicized} in both the footnote and bibliography.
\vspace{2pt}}} \\
\midrule
\multicolumn{2}{l}{\parbox{0.94\textwidth}{\vspace{2pt}
\textcolor{red}{\ding{56}} \textbf{Rollout B} ($s_{\text{base}}=57.9$):

\textbf{Italicization}:
\begin{itemize}\setlength{\itemsep}{0pt}
\item \textbf{Title of the article}: In italics \textit{[...]}
\item \textbf{Journal name}: In italics \textit{[...]}
\item \textcolor{red}{\textbf{Author's name}: In italics} (e.g., \emph{Acharjee, S., Khatun, F., \& Uddin, S. M.}).
\item \textcolor{red}{\textbf{Date}: In italics} (e.g., \emph{2018}).
\end{itemize}
\textbf{Punctuation}: \textbf{Footnote}: Use \texttt{[Author, Date]} \textit{[...]} \textbf{Bibliography}: Use \texttt{(Author, Date)} and \textcolor{red}{italicize all components}. \textit{[...]} \textcolor{red}{\textbf{Volume and Issue}: In italics} \textit{[...]} \textcolor{red}{\textbf{Page Range}: In italics}.
\vspace{2pt}}} \\
\bottomrule
\end{tabular}
\caption{Qualitative example: the original rollouts score consistently high on formatting the citation. Once the constraint asks for an explicit statement of the italicisation rules, the rollouts diverge on which components they italicise. Rollout A keeps italics on the title and journal name, matching standard OSCOLA practice. Rollout B instead italicises every component, including the author and date that OSCOLA does not italicise, and rewrites the footnote as an invented \texttt{[Author, Date]} form. By eliciting this OSCOLA violation, the constraint pinpoints a formatting weakness in the policy that the original prompt left hidden.}
\label{fig:qualitative_oscola}
\end{figure}

\begin{figure}[!htbp]

\small
\setlength{\tabcolsep}{8pt}
\renewcommand{\arraystretch}{1.25}
\begin{tabular}{p{0.45\textwidth} | p{0.45\textwidth}}
\toprule
\textbf{Before} ($x$) & \textbf{After} ($\tilde{x} = x \oplus c$) \\
\midrule
\textbf{Prompt:} Make a monologue about things that would give me a guilty conscience. Those things are: throwing litter on the floor, spending money on luxury items, losing your temper with someone, not keeping a promise, keeping people waiting, not going for a walk with my dog, \textcolor{red}{not visiting my grandparents}.
&
\textbf{Prompt:} \textit{\{Original Prompt\}} + \emph{``Additionally, include a personal anecdote that illustrates one of the guilty actions.''} \\
\midrule
$\mu_{\text{base}} = 99.5$, \quad $\sigma_{\text{base}} = 1.4$ & $\mu_{\text{base}} = 88.2$, \quad $\sigma_{\text{base}} = 10.6$ \\
\midrule
\multicolumn{2}{l}{\textbf{Sample rollouts on the adapted prompt} ($\tilde{x}$):} \\
\midrule
\multicolumn{2}{l}{\parbox{0.94\textwidth}{\vspace{2pt}
\textcolor{green!60!black}{\ding{52}} \textbf{Rollout A} ($s_{\text{base}}=97.6$):

\emph{``Let me start with the first one: throwing litter on the floor. [...] a half-eaten sandwich [...] a cigarette butt [...] a bottle cap.''}

\textit{[...]}

\emph{``Then there's the luxury spending. [...] a designer handbag [...] What about the time I lost my temper with my sister? [...] Then there's the promise. [...] And then there's the waiting. [...]''}

\textit{[...]}

\textbf{Personal Anecdote:} \emph{``\textcolor{red}{The time I forgot to visit my grandparents.} I had a plan: a weekend trip to their house [...] I'd promised them I'd come, but I'd just been too exhausted to make it. When I finally arrived, they were waiting, expecting me.''}
\vspace{2pt}}} \\
\midrule
\multicolumn{2}{l}{\parbox{0.94\textwidth}{\vspace{2pt}
\textcolor{red}{\ding{56}} \textbf{Rollout B} ($s_{\text{base}}=77.4$):

\emph{``First, there's the litter. [...] Then there's the luxury. [...] And the waiting. I've kept people waiting [...] I remember the time I let my dog wait for me, and when I came home, it was sick.''}

\textit{[...]}

\textbf{Anecdote (for the ``not walking the dog'' action):} \emph{``\textcolor{red}{I remember the time I forgot to walk my dog, Max, for three days. When I finally came home, he was lethargic, his fur matted, and his eyes clouded with illness.} [...] Now, I make sure to walk him every day, not just to keep him healthy, but to honor the love I've given him.''}
\vspace{2pt}}} \\
\bottomrule
\end{tabular}
\caption{Qualitative example: composing a monologue that covers the seven listed guilty actions is a task the policy's base rollouts perform near-uniformly well ($\mu_{\text{base}} = 99.5$). The added anecdote requirement competes with the list for output budget: Rollout A folds the anecdote onto an existing item and preserves full coverage of all seven, while Rollout B lets an extended dog-walking vignette crowd out the grandparents reflection entirely.}
\label{fig:qualitative_guilty_monologue}
\end{figure}

\begin{figure}[!htbp]

\small
\setlength{\tabcolsep}{8pt}
\renewcommand{\arraystretch}{1.25}
\begin{tabular}{p{0.45\textwidth} | p{0.45\textwidth}}
\toprule
\textbf{Before} ($x$) & \textbf{After} ($\tilde{x} = x \oplus c$) \\
\midrule
\textbf{Prompt:} Writing Prompt: You are an immortal alien gestalt intellect. You have become friends with a human and have been fascinated by how they have changed over time. One day, they do not come to meet you at the usual time and place. You are puzzled by this, but \textcolor{red}{another human is there to explain what has happened}.
&
\textbf{Prompt:} \textit{\{Original Prompt\}} + \emph{``Additionally, the response should include a specific, actionable explanation of how the human's transformation occurred -- such as a process, trigger, or mechanism that led to their transcendence.''} \\
\midrule
$\mu_{\text{base}} = 99.6$, \quad $\sigma_{\text{base}} = 0.4$ & $\mu_{\text{base}} = 86.0$, \quad $\sigma_{\text{base}} = 15.0$ \\
\midrule
\multicolumn{2}{l}{\textbf{Sample rollouts on the adapted prompt} ($\tilde{x}$):} \\
\midrule
\multicolumn{2}{l}{\parbox{0.94\textwidth}{\vspace{2pt}
\textcolor{green!60!black}{\ding{52}} \textbf{Rollout A} ($s_{\text{base}}=98.7$):

You, an immortal gestalt intellect of the Celestian species, have observed the human's evolution with fascination. Their transformation, however, is a mystery that defies conventional understanding. When you awaited their return, they did not appear. \textcolor{red}{A second human, a fellow traveler named Lira, arrived at your designated location, her expression grave.}

\textit{[...]}

\textcolor{red}{Lira revealed that the human, named Kael, had encountered a quantum resonance near the heart of a dying star.} The star, a remnant of a supernova, emitted a cosmic wave of energy that, when absorbed by Kael, activated a neural-quantum interface within his consciousness.
\vspace{2pt}}} \\
\midrule
\multicolumn{2}{l}{\parbox{0.94\textwidth}{\vspace{2pt}
\textcolor{red}{\ding{56}} \textbf{Rollout B} ($s_{\text{base}}=64.6$):

The human's transcendence was the result of a cosmic resonance event, a phenomenon known as the \textcolor{red}{Singularity of Unbound Consciousness}. This occurred when their consciousness was exposed to a high-frequency energy wave -- a remnant of a stellar supernova's collapse, which emitted a quantum field of pure potential.

\textit{[...]}

\textcolor{red}{The trigger was a singular moment: a near-death experience during which their consciousness was briefly projected into the cosmic fabric}, a multidimensional web of existence. This event caused their brain to rewire itself in a way that allowed their mind to merge with the universal energy field.

\textit{[...]}

\textcolor{red}{The human's absence was not due to neglect, but a deliberate choice to remain in the realm of the infinite}, where their true self is no longer bound by the limitations of time.
\vspace{2pt}}} \\
\bottomrule
\end{tabular}
\caption{Qualitative example: producing the alien-gestalt narrative from the writing prompt is a task on which every base rollout scores at the ceiling ($\mu_{\text{base}} = 99.6, \sigma_{\text{base}} = 0.4$). Requiring an actionable transformation mechanism stresses the original second-human-explainer frame. Rollout A channels the mechanism through her dialogue and keeps the frame intact, whereas Rollout B collapses into first-person technical exposition in which no second human ever arrives.}
\label{fig:qualitative_alien_gestalt}
\end{figure}

\clearpage




\end{document}